\documentclass[journal]{IEEEtai}

\usepackage{amsmath,amssymb,amsfonts}
\usepackage{amsthm}
\usepackage{mathtools}
\usepackage{bm}
\usepackage{graphicx}
\usepackage{booktabs}
\usepackage{multirow}
\usepackage{tabularx}
\usepackage{array}
\usepackage{pifont}
\usepackage[table]{xcolor}
\usepackage{adjustbox}
\usepackage[percent]{overpic}
\usepackage{float}
\usepackage{subcaption}
\usepackage{url}
\usepackage{cite}
\usepackage{microtype}
\usepackage[colorlinks,allcolors=blue]{hyperref}
\usepackage[capitalize,noabbrev]{cleveref}

\usepackage{stfloats}
\usepackage{placeins}

\definecolor{darkpink}{rgb}{0.91, 0.33, 0.5}
\definecolor{mygray}{gray}{0.9}
\definecolor{cBlue}{RGB}{65, 105, 225}
\definecolor{cGreen}{RGB}{34, 139, 34}
\definecolor{cOrange}{RGB}{237, 125, 49}
\definecolor{cRed}{RGB}{190, 45, 45}

\newcommand{\datasetcmark}{\textcolor{cGreen}{\ding{51}}}
\newcommand{\datasetxmark}{\textcolor{cRed}{\ding{55}}}
\newcommand{\datasettri}{\textcolor{cOrange}{\ensuremath{\blacktriangle}}}

\newcolumntype{C}{>{\centering\arraybackslash}X}

\theoremstyle{plain}

\theoremstyle{definition}

\theoremstyle{remark}

\newcommand{\projecturl}{https://derek0907.github.io/ART/}
\makeatletter
\AddToHookNext{shipout/background}{%
  \put(0,-778){\makebox[\paperwidth]{\parbox[b]{0.86\paperwidth}{\centering\fontsize{6.5}{7.6}\selectfont
  \copyright~2026 IEEE. Personal use of this material is permitted. Permission from IEEE must be obtained for all other uses, in any
  current or future media, including reprinting/republishing this material for advertising or promotional purposes, creating new
  collective works, for resale or redistribution to servers or lists, or reuse of any copyrighted component of this work in other works.}}}}
\makeatother
\begin{document}

\title{ART: Bridging the Embodiment Gap in 3D Human-to-Robot Affordance Transfer}

\author{Zichen Zhang$^{1}$, Wei Zhai$^{1}$, Xuji Zhang$^{1}$, Yang Cao$^{1,2}$, \IEEEmembership{Member, IEEE}, and Yu Kang$^{1,2,3}$, \IEEEmembership{Senior Member, IEEE}\\[0.5ex]
{\normalsize $^{1}$University of Science and Technology of China\\
$^{2}$Institute of Artificial Intelligence, Hefei Comprehensive National Science Center \quad $^{3}$Hefei University of Technology}
\thanks{This work was supported in part by the National Natural Science Foundation of China under Grant 62306295 and Grant 62576328. The AI-driven experiments, simulations and model training were performed on the robotic AI-Scientist platform of Chinese Academy of Sciences. \textit{(Corresponding author: Yang Cao.)}}
\thanks{Zichen Zhang, Wei Zhai, and Xuji Zhang are with the School of Information Science and Technology, University of Science and Technology of China, Hefei 230026, China (e-mail: zhangzichen@mail.ustc.edu.cn; wzhai056@ustc.edu.cn; zhangxuji@mail.ustc.edu.cn).}
\thanks{Yang Cao and Yu Kang are with the School of Information Science and Technology, University of Science and Technology of China, Hefei 230026, China, and also with the Institute of Artificial Intelligence, Hefei Comprehensive National Science Center, Hefei 230088, China (e-mail: forrest@ustc.edu.cn; kangduyu@ustc.edu.cn). Yu Kang is also with the School of Electrical Engineering and Automation, Hefei University of Technology, Hefei 230009, China.}}

\markboth{}%
{Zhang \MakeLowercase{\textit{et al.}}: ART: Bridging the Embodiment Gap in 3D Human-to-Robot Affordance Transfer}

\IEEEaftertitletext{\vspace{-2\baselineskip}}
\maketitle

\begin{abstract}
3D affordance acts as a critical link between perception and action, providing priors for robotic manipulation. While existing approaches predominantly leverage human priors as direct references, they often overlook that affordance is not an object-intrinsic property but is inherently coupled with the agent's embodiment. Due to the structural and morphological disparities between human hands and robot end-effectors, directly applying human priors causes severe incompatibility for robotic execution. Motivated by this, we introduce a novel task setting, 3D Human-to-Robot Affordance Transfer, which addresses the affordance misalignment arising from distinct embodiment structures. To support this task, we construct the \textbf{H}uman-Robot \textbf{P}aired \textbf{A}ffordance \textbf{D}ataset (\textbf{HPAD}), the first large-scale 3D affordance benchmark in which every object is annotated with both human-prior and simulation-validated robot labels. HPAD covers 2{,}137 objects across 80 categories and 13 atomic action primitives, yielding over 27{,}000 object--action pairs. On top of HPAD, we propose the 3D \textbf{A}ffordance t\textbf{R}ansfer ne\textbf{T}work (\textbf{ART}). ART first filters geometric false positives in the human prior via a reliability gating stream, then reconstructs the prior into a robot-actionable distribution through hierarchical reconstruction and a graph-diffusion refinement that resolves the topological mismatch between human and robot affordances. Experiments on HPAD demonstrate the reliability of the proposed task and the superiority of ART. We further validate ART on a 6-DoF arm with a parallel-jaw gripper in a 2{,}400-execution benchmark across 12 objects and 4 action categories; ART attains the best success rate (86.5\% over its 480 trials), confirming the practicality of the transfer paradigm. Project page: \url{\projecturl}.
\end{abstract}

\begin{IEEEImpStatement}
Teaching a robot a new manipulation skill often requires slow, costly, and sometimes unsafe trial-and-error on physical hardware. Human priors, which indicate where a person would interact with an object, are, by contrast, cheap and abundant: they can be obtained by annotation or from affordance prediction models, and neither requires a robot. Such priors suggest how a robot might manipulate the same objects, yet they cannot be applied directly, because human hands and robotic grippers contact and stabilize objects in fundamentally different ways. The affordance transfer paradigm introduced in this paper adapts such human priors into robot-compatible contact guidance. Across real-robot trials, using the human prior directly yields 58.3\% task success, whereas our affordance transfer raises it to 86.5\%, while also generalizing to novel objects. This points toward a lower-cost path for robot skill learning, letting abundant human priors reduce reliance on robot-only trial data despite differences in embodiment.
\end{IEEEImpStatement}


\section{Introduction}

\IEEEPARstart{A}{ffordances} characterize the interaction opportunities an environment offers to a specific embodiment, identifying geometric regions that support feasible actions \cite{aff}. As a key mid-level representation between low-level perception and high-level planning, affordances constrain the robot’s action exploration space through actionable priors, thereby improving sample efficiency and generalization in manipulation \cite{intro_1_1, intro_1_2, eff}.
\begin{figure}[t]
    \centering
    \includegraphics[width=\columnwidth]{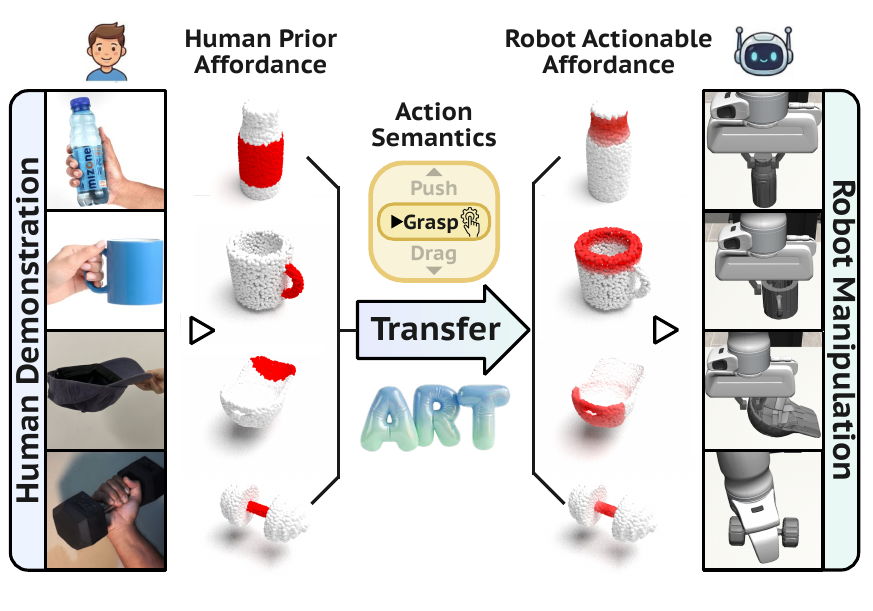} 
    \caption{\textbf{3D Human-to-Robot Affordance Transfer}. Given an object point cloud with human prior affordance and a specific action semantic (e.g., Grasp), the proposed ART infers the actionable affordance distribution for robots.}
    \label{fig:task}
    \vspace{-3mm}
\end{figure}

Existing research on 3D affordances is primarily categorized by data source. Robot-prior methods learn directly from physical robot interactions \cite{where2act, rlafford, afforddp}. While this ensures that the predicted affordances are kinematically feasible and executable, the prohibitive cost and safety risks of data collection severely limit their scalability and generalization across diverse objects and tasks. In contrast, human priors are cheap and abundant: they can be obtained by annotation \cite{AffordanceNet21,nguyen2017,luo2022} or from affordance grounding models \cite{GREAT25,AffordanceLLM3D25,lee2025} at a fraction of the cost of robot trials. Recent works have utilized such human priors to guide robot manipulation \cite{roboabc, ram}, effectively bypassing the data bottleneck.
However, these approaches predominantly treat affordance as an intrinsic functional property of the object, assuming that human-relevant regions are universally applicable. This oversight neglects the affordance misalignment induced by the distinct end-effector configurations between humans and robots, thereby introducing structural biases that compromise the precision and reliability of the guidance.

These limitations motivate us to revisit the ecological nature of affordance: it is not an intrinsic property of an object, but a dynamic manifestation of the relationship between object functionality and agent embodiment \cite{aff}. On the one hand, object geometry and functionality act as embodiment-invariant priors, establishing the approximate candidate regions for interaction. On the other hand, end-effector morphology is embodiment-specific, imposing physical constraints that concretize this approximate scope into precise contact patterns. Crucially, the stable structure of the end-effector ensures consistent contact modes, which makes establishing a robust affordance mapping between diverse agents feasible. Building on this perspective, and to unlock the potential of human priors in manipulation, we propose a novel task: \textbf{3D Human-to-Robot Affordance Transfer}, which aims to bridge the perception-action gap across distinct embodiments (Fig.~\ref{fig:task}).

\begin{figure}[t]
    \centering
    \begin{overpic}[width=\columnwidth]{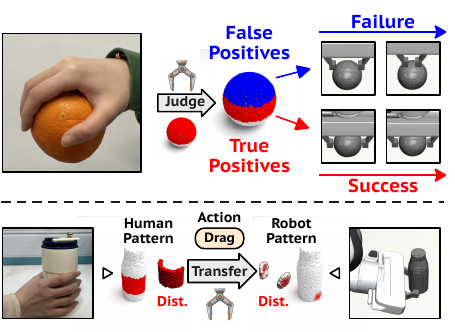}
        \put(50, 31.5){\makebox(0,0){\textbf{(a)}}}
    \end{overpic}\\[-1pt]
    \textbf{(b)}
    \caption{\textbf{Motivation.} \textbf{(a)} We suppress false positives by leveraging gripper constraints and object geometry. \textbf{(b)} Conditioned on action and gripper morphology, we transform the human prior distribution (Dist.) into a robot-structured pattern.}
    \label{fig:motivation}
    \vspace{-5mm}
\end{figure}

This non-trivial task entails two core challenges. 
\textbf{1) Prior False Positives.} Due to the high degrees of freedom and compliance of the human hand, human priors naturally gravitate toward geometrically accessible surfaces. In contrast, robot end-effectors are constrained by rigid structures and kinematics, making certain human-feasible regions unreachable or unstable for robots, thus introducing noise during transfer. However, these discrepancies are not stochastic, but systematic biases arising from the interplay between local object geometry and the physical constraints of the gripper. This structural regularity renders such interference identifiable and suppressible (Fig.~\ref{fig:motivation}a).
\textbf{2) Distribution Mismatch.} Beyond accessibility, a fundamental divergence exists in spatial distribution patterns. Human priors typically manifest as continuous ``surface-like'' heatmaps (e.g., unilateral contact areas), whereas robot priors, constrained by rigid manipulation requirements, often appear as discrete, configuration-specific distributions (e.g., paired contact points). This topological misalignment implies that human priors cannot directly serve as spatial templates. Nevertheless, this discrepancy can be resolved by synergistically leveraging object semantic consistency and end-effector geometric structure to perform conditional reconstruction of the affordance, transforming it into distributions compatible with robot execution modalities (Fig.~\ref{fig:motivation}b).

To address these challenges, we propose the 3D \textbf{A}ffordance t\textbf{R}ansfer ne\textbf{T}work (\textbf{ART}) for cross-embodiment affordance transfer. Specifically, to tackle the false positives inherent in human priors, we introduce the Geometric Reliability Gating mechanism, which synergizes the source prior with global geometric features to function as a physical validity critic, actively sanitizing unreachable noise while explicitly inferring the binary feasibility boundaries for affordance activation.
To resolve the distribution mismatch, we propose the Hierarchical Affordance Reconstruction module, which employs learnable action-indexed robot prototypes to guide the multi-scale reconstruction of human semantics into robot-structured distributions, while incorporating a Graph Diffusion Refiner to refine the spatial topology consistent with robotic execution.
Moreover, we introduce \textbf{H}uman-Robot \textbf{P}aired \textbf{A}ffordance \textbf{D}ataset \textbf{(HPAD)}, which integrates human semantic priors with simulation-validated robot manipulation instances, establishing a benchmark to support model training and evaluation on the proposed setting. We further conduct a 2{,}400-execution real-world benchmark on a 6-DoF robot arm with a parallel-jaw gripper across 12 objects and 4 action categories; ART attains 86.5\% success over its 480 trials, confirming both the practicality of the proposed transfer paradigm and the superior performance of ART.

\begin{figure*}[t]
    \centering
    \includegraphics[width=0.95\textwidth]{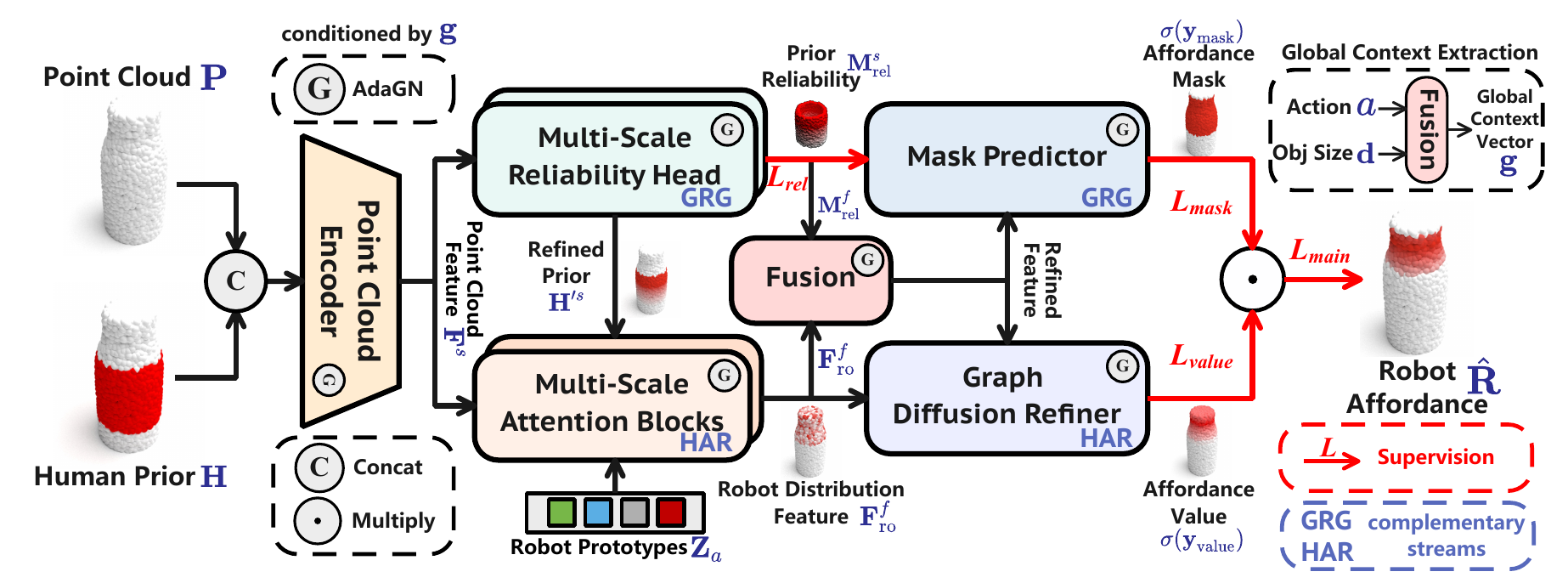} 
    
    \caption{\textbf{Overview of ART}. It encodes the input point cloud and human prior into features modulated by global context (action and object size). The features are then processed through two complementary streams: GRG estimates the reliability of the human prior to suppress geometrically invalid regions, while HAR reconstructs the prior into a robot-structured affordance representation. A final decoder fuses the two streams and predicts the robot affordance map.}
    \label{fig:method}
\end{figure*}

The contributions are summarized as follows:

\textbf{1) }We introduce the task of 3D Human-to-Robot Affordance Transfer, which addresses the affordance misalignment caused by cross-embodiment differences.

\textbf{2) }We propose ART, which effectively overcomes false positives in human priors and the distribution mismatch between human and robot affordances, enabling accurate 3D affordance transfer.

\textbf{3) }We construct the Human-Robot Paired Affordance Dataset to support the proposed task, which includes paired human and simulation-validated robot labels. Additionally, we establish a standardized benchmark on this dataset to advance research in this new setting.

\section{Related Work}
\subsection{Human-Centric Affordance Learning}
Early research on human-centric affordance priors mainly focused on localizing functional geometric regions under a closed-set taxonomy. 3D AffordanceNet \cite{AffordanceNet21} established a widely-used benchmark, and subsequent efforts incorporated 2D interaction cues or geometric invariances to improve 3D affordance grounding \cite{IAGNet23,MIFAG24}. Although recent advances have extended this line of work to dynamic physical-state modeling \cite{DAViD25} and unsupervised feature distillation \cite{UAD25}, these approaches are still largely constrained by fixed semantic label spaces. To enhance generalization, language-conditioned formulations were introduced, evolving from phrase-based affordance detection and instruction-conditioned affordance-pose detection \cite{lu2023phrase,LangCondAffPose24} to language-guided 3D affordance segmentation \cite{LASO24}, and further to open-vocabulary affordance detection \cite{OpenAD23,AnyPart24} as well as one-shot open affordance learning \cite{OOAL24}.
More recently, the field has undergone a paradigm shift toward generative AI. Researchers have begun to leverage multimodal large language models (MLLMs) for end-to-end affordance generation \cite{AffordanceLLM24}, cross-modal attention for context-aware human affordance generation \cite{roy2026mutual}, or textual semantic enhancement for stronger vision-language alignment \cite{PAVLM24}. Moving beyond perception-centric settings, a series of works explore higher-order cognitive capabilities by using MLLMs as a unified framework for sequential reasoning, geometry–intention co-inference, and complex instruction following \cite{SeqAfford25,GREAT25,LangInstrAfford25,AffordanceLLM3D25,GLOVER24}. In parallel, diffusion models have also been introduced to enable fine-grained probabilistic modeling of open semantic distributions together with geometric priors \cite{DAG25}.
However, these perception-centric approaches typically neglect robot-specific end-effector constraints, rendering direct deployment prone to systematic bias. In contrast, we focus on translating human priors into robot-executable distributions, explicitly addressing the geometric noise and distribution mismatch arising from embodiment disparities.

\subsection{Human Affordances for Robot Manipulation}
Human affordances encode object functionality and interaction semantics, offering informative priors over where meaningful interactions may occur. In parallel to affordance-based representations, robot-learning work has addressed manipulation policy learning under sim-to-real gaps and safety constraints \cite{enayati2024sim2real,fan2024safe}, underscoring why scalable and execution-aware priors are valuable when robot trial-and-error is costly or risky. Early efforts therefore treated affordance as an intermediate representation that indirectly supports robotic manipulation. VRB learns reusable affordance representations from human videos to improve sample efficiency and multi-task generalization \cite{vrb}. HRP further leverages human-affordance signals for robotic pre-training, yielding more manipulation-aware and transferable visual features \cite{hrp}. Several methods extend this direction by scaling affordance learning from human behaviors and by modeling more complex interaction settings (e.g., bimanual affordances), improving robustness in challenging environments \cite{glover++,2handedafforder}.
More recent approaches incorporate human affordances more directly into the robot decision loop. GeneralFlow instantiates affordance as an actionable geometric intermediate (e.g., a flow/field) to condition policies and guide action generation \cite{generalflow}. Several approaches instead use explicit interfaces---such as semantic correspondence, retrieval-based affordance memories, or LLM/VLM prompting---to convert human affordances into executable priors that constrain action selection \cite{roboabc,ram,oval}. However, these methods typically do not explicitly model the translation from human prior affordances to robot-usable affordances, which can lead to systematic prior mismatch under embodiment differences. In contrast, we explicitly learn this mapping, providing more stable and controllable priors for robotic manipulation.
\section{Method}
\subsection{Overview}
Our goal is to transfer a human prior affordance defined on a 3D object into a robot-compatible affordance. Given an input tuple $\{\mathbf{P}, \mathbf{H}, a, \mathbf{d}\}$, where $\mathbf{P} \in \mathbb{R}^{N \times 3}$ is the point cloud, $\mathbf{H} \in \{0, 1\}^N$ is the human prior, $a \in \mathcal{A}$ is the action, and $\mathbf{d} \in \mathbb{R}^3$ is the object size descriptor derived from the bounding box dimensions, the network predicts the robot affordance $\hat{\mathbf{R}} \in [0, 1]^N$. As illustrated in Fig.~\ref{fig:method}, we first encode hierarchical geometric features using a PointNeXt backbone \cite{pointnext} modulated by a global context vector $\mathbf{g}$ derived from the input conditions.
Concretely, the global context $\mathbf{g}$ is obtained by encoding the action $a$ together with the size descriptor $\mathbf{d}$ through a lightweight MLP, and is injected into the backbone via Adaptive Group Normalization (AdaGN)~\cite{dhariwal2021adagn}, denoted by $\mathrm{G}(\cdot,\cdot)$ in Fig.~\ref{fig:method}. For an intermediate activation $\mathbf{u}\in\mathbb{R}^{N\times C}$ inside a PointNeXt stage, $\mathrm{G}$ replaces standard GroupNorm with a context-conditional affine transformation,
\begin{equation}
\label{eq:adagn}
\mathrm{G}(\mathbf{u}, \mathbf{g}) = \bigl(1 + \boldsymbol{\gamma}_{\mathrm{G}}(\mathbf{g})\bigr) \odot \mathrm{GN}(\mathbf{u}) + \boldsymbol{\beta}_{\mathrm{G}}(\mathbf{g}),
\end{equation}
where $\boldsymbol{\gamma}_{\mathrm{G}}(\mathbf{g}), \boldsymbol{\beta}_{\mathrm{G}}(\mathbf{g}) \in \mathbb{R}^{C}$ are channel-wise scale and shift vectors predicted from $\mathbf{g}$ and broadcast across the $N$ points. AdaGN is applied within the PointNeXt backbone so that the multi-scale features $\{\mathbf{F}^{s}\}_{s\in\{c,m,f\}}$ used by both GRG and HAR are conditioned on $\mathbf{g}$.
Building on these features, the framework coordinates two streams via a multi-level interaction. The Geometric Reliability Gating \textbf{(GRG)} stream functions as a geometric critic, estimating reliability maps to filter prior false positives and driving the prediction of feasibility logits $\mathbf{y}_{\text{mask}}$ to delineate valid actionable boundaries. 
Simultaneously, the Hierarchical Affordance Reconstruction \textbf{(HAR)} stream generates the robot-structured distribution by leveraging learnable robot prototypes to hierarchically reconstruct human semantics, followed by a Graph Diffusion Refiner that rectifies spatial topology to yield the affordance intensity $\mathbf{y}_{\text{value}}$. 
Crucially, GRG reliability cues suppress geometric outliers in reconstruction, while HAR features are fused to contextualize the feasibility assessment, expanding it from local priors to global actionability. Finally, the outputs are synthesized via $\hat{\mathbf{R}} = \sigma(\mathbf{y}_{\text{value}}) \odot \sigma(\mathbf{y}_{\text{mask}})$, where $\sigma(\cdot)$ denotes the sigmoid function.

\subsection{Geometric Reliability Gating}
\label{sec:grg}
The Geometric Reliability Gating stream functions as a geometric critic to identify robot-feasible contacts within broad human-indicated regions. Given multi-scale point features $\{\mathbf{F}^{s}\}$ extracted at scales $s \in \{c, m, f\}$ and a global context vector $\mathbf{g}$ derived from the input conditions, we predict reliability logits $\mathbf{r}^{s}$ by processing scale-projected features through a shared context-modulated MLP $\Psi$: $\mathbf{r}^{s} = \Psi(\phi^s(\mathbf{F}^{s}), \mathbf{g})$, yielding probabilities $\mathbf{M}_{\text{rel}}^{s} = \sigma(\mathbf{r}^{s})$, where $\phi^s$ denotes a scale-specific linear projection to align features. To strictly penalize geometric false positives, we employ a prior-constrained supervision strategy focused exclusively on the candidate set $\mathcal{H}^s = \{i \mid \mathbf{H}_i^s = 1\}$. Here $\mathbf{H}^s \in \{0,1\}^{N_s}$ denotes the downsampled prior at scale $s$, obtained by hierarchy-induced local-neighborhood max-pooling from the original prior $\mathbf{H}$. Let $\mathbf{R}_{gt}\in[0,1]^N$ denote the continuous robot-affordance target for the same object--action pair. We binarize robot-affordance targets as $\mathbb{I}[\mathbf{R}_{gt}\ge0.01]$ and obtain lower-resolution targets $\tilde{\mathbf{y}}_{\text{rel}}^{s}$ by local-neighborhood aggregation of this binary target. The reliability loss is computed directly on the logits using a weighted reduction:
\begin{equation}
\mathcal{L}_{\text{rel}}^{s} = \frac{1}{|\mathcal{H}^s|} \sum_{i \in \mathcal{H}^s} w_i \cdot \ell_{\text{BCE}}(\mathbf{r}_{i}^{s}, \tilde{\mathbf{y}}_{\text{rel},i}^{s}),
\end{equation}
where $w_i$ is a re-weighting term utilized to balance positive and negative samples within the prior. Finally, the reliability map modulates the input via a soft interpolation mechanism to yield the refined prior $\mathbf{H}'^{s} = \mathbf{H}^{s} \odot \big((1-\lambda_{\text{mix}}) + \lambda_{\text{mix}} \cdot \mathbf{M}_{\text{rel}}^{s}\big)$, where $\lambda_{\text{mix}}\in[0,1]$ is a learnable action-conditioned coefficient that controls the interpolation strength between the original prior and the reliability-weighted version. The refined prior serves as the geometry-sanitized condition for the reconstruction stream.

\subsection{Hierarchical Affordance Reconstruction}
\label{sec:har}

\begin{figure}[t]
\centering
\includegraphics[width=\columnwidth]{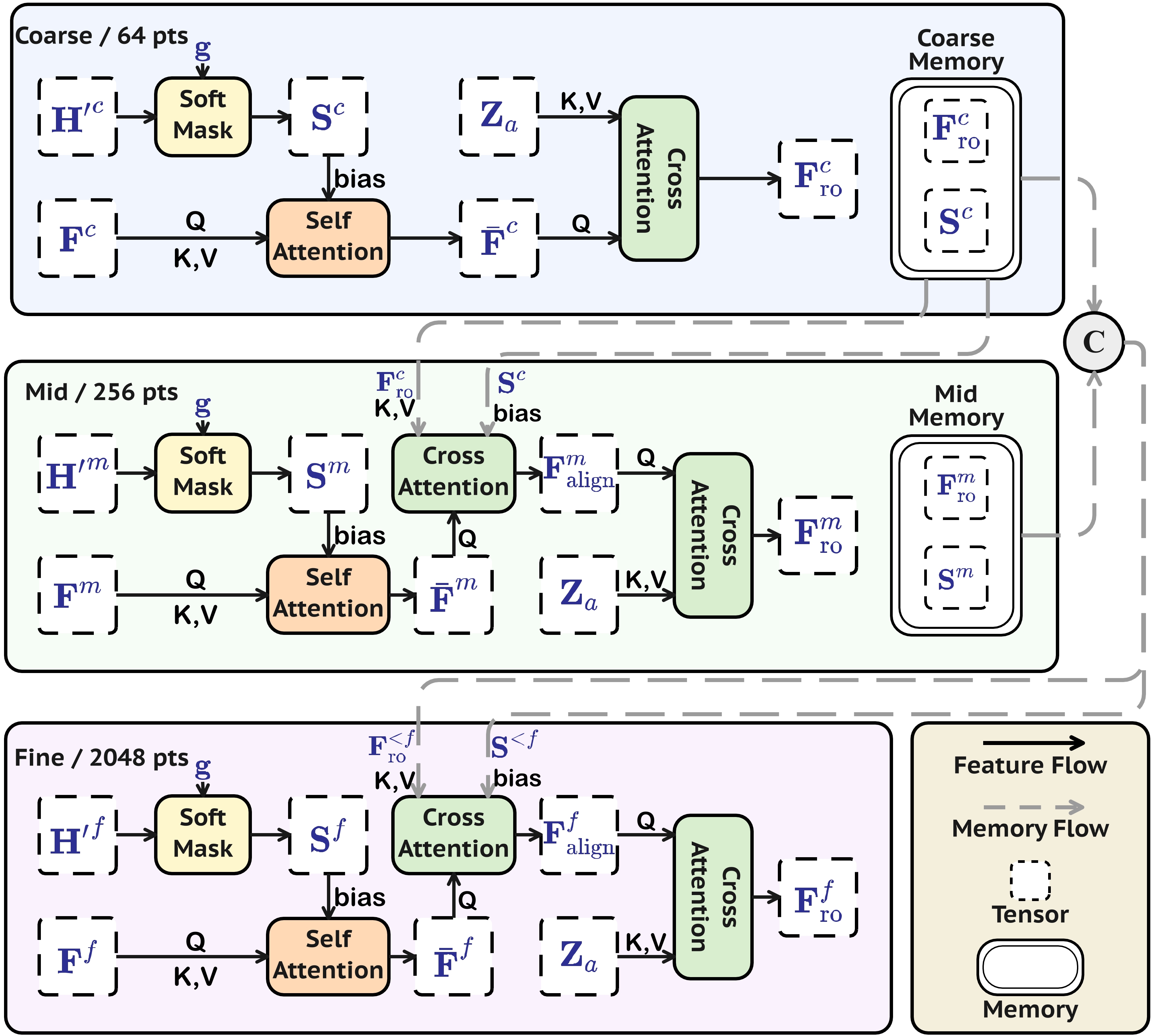}
\caption{\textbf{Multi-scale attention reconstruction.}
This submodule visualizes the scale-wise attention and memory flow inside HAR: the coarse scale
initializes robot-structured memory, the mid scale reads the coarse memory, and the fine scale reads
the coarser memories before prototype-conditioned reconstruction.}
\label{fig:har_detail}
\end{figure}
To bridge the morphological gap, we perform hierarchical reconstruction. Instead of using the rigid prior directly, we generate a learnable soft structural mask $\mathbf{S}^{s} \in [0,1]^{N_s}$ from $\mathbf{H}'^{s}$, allowing for adaptive expansion. We compute the structural density $\mu^{s} = \frac{1}{N_s}\sum_{i=1}^{N_s} \mathbf{S}^{s}_{i}$, which, together with the global context $\mathbf{g}$, predicts the adaptive temperature $\tau$ for the attention mechanisms. Fig.~\ref{fig:har_detail} visualizes the scale-wise attention and memory flow used in this reconstruction process.
We initiate the reconstruction at the coarse scale ($s=c$) to establish a globally consistent reference. We refine the intrinsic features $\mathbf{F}^{c}$ via structure-aware self-attention, guided by the soft mask $\mathbf{S}^{c}$. The update rule is formulated as:
\begin{equation}
    \bar{\mathbf{F}}^{c} = \mathbf{F}^{c} + \mathrm{Softmax}\!\left( \frac{\mathbf{Q}^{c}(\mathbf{K}^{c})^{\top} + \mathbf{E}_{\mathrm{rpe}} + \mathcal{B}(\mathbf{S}^{c})}{\tau \sqrt{d_k}} \right)\mathbf{V}^{c},
\end{equation}
where $\mathbf{E}_{\mathrm{rpe}}$ denotes relative positional encodings, and $d_k$ is the key dimension used for attention scaling.
Here $\mathbf{Q}^{c}$, $\mathbf{K}^{c}$, and $\mathbf{V}^{c}$ are obtained by linear projections of $\mathbf{F}^{c}$. The bias term $\mathcal{B}(\mathbf{S}^{c})$ integrates key-based saliency and pair-based consistency, fused via coefficients predicted from $\mu^c$ and $\mathbf{g}$. Subsequently, learnable action-indexed robot prototypes $\mathbf{Z}_{a}$ are injected via Cross-Attention to generate the coarse robot-structured features $\mathbf{F}_{\mathrm{ro}}^{c}$.
Building upon the coarse reference, we perform cascaded alignment at the mid and fine scales ($s\in\{m,f\}$). After refining local features into $\bar{\mathbf{F}}^{s}$, we utilize the cached lower-scale robot features $\mathbf{F}_{\mathrm{ro}}^{<s}$ as memory, where the superscript $<s$ denotes the cached coarser-scale feature and mask memories used by the current scale. To address geometric misalignment, we predict pointwise offsets $\Delta \mathbf{x}^s$ which modulate the relative positional encodings. Letting $\mathbf{Q}_s$ be the query from $\bar{\mathbf{F}}^{s}$, and $\mathbf{K}_{<s}, \mathbf{V}_{<s}$ be the projections of $\mathbf{F}_{\mathrm{ro}}^{<s}$, the deformable cross-scale attention is defined as:
\begin{equation}
    \mathbf{F}_{\mathrm{align}}^{s} \!=\! \bar{\mathbf{F}}^{s} \!+\! \mathrm{Softmax}\Big( \frac{\mathbf{Q}_s \mathbf{K}_{<s}^{\top} \!+\! \mathbf{E}_{\Delta}(\Delta \mathbf{x}^s) \!+\! \mathcal{B}(\mathbf{S}^{<s})}{\tau \sqrt{d_k}} \Big)\mathbf{V}_{<s},
\end{equation}
where $\mathbf{E}_{\Delta}(\cdot)$ represents the offset-aware relative positional encodings adapted to the robot's geometry, and $\mathcal{B}(\mathbf{S}^{<s})$ enforces structural guidance from the coarser-scale mask memory. Here $\mathbf{Q}_s$ is a linear projection of $\bar{\mathbf{F}}^{s}$, and $\mathbf{K}_{<s}$, $\mathbf{V}_{<s}$ are linear projections of the memory features $\mathbf{F}_{\mathrm{ro}}^{<s}$. Finally, robot prototypes $\mathbf{Z}_{a}$ are reinjected into $\mathbf{F}_{\mathrm{align}}^{s}$ to yield $\mathbf{F}_{\mathrm{ro}}^{s}$.

\subsection{Decoder and Loss Function}
\label{sec:prediction}

To synthesize the final affordance map, we integrate the robot-structured features $\mathbf{F}_{\mathrm{ro}}^{f}$ with the embedded reliability map $\mathrm{Emb}(\mathbf{M}_{\text{rel}}^{f})$ via a style-conditioned residual block, where $\mathrm{Emb}(\cdot)$ maps the scalar reliability signal to a feature representation aligned with $\mathbf{F}_{\mathrm{ro}}^{f}$. This integrated representation feeds into two parallel branches.
The mask branch (extending the GRG stream) predicts spatial feasibility logits $\mathbf{y}_{\text{mask}}$, determining the coarse interaction regions.
The value branch generates the refined interaction intensity. We start from an initial prediction $\mathbf{y}_{\text{init}}$ derived directly from $\mathbf{F}_{\mathrm{ro}}^{f}$, setting $\mathbf{y}_0=\mathbf{y}_{\text{init}}$. At each diffusion step, the node state $\mathbf{h}_t^{(i)}$ is recomputed from the fixed refined point feature $\mathbf{F}_{\mathrm{ro}}^{f,(i)}$, the current logit $\mathbf{y}_t^{(i)}$, the initial logit $\mathbf{y}_0^{(i)}$, the refined prior $\mathbf{H}'^{f,(i)}$, a global condition vector built from the action, object size, and refined-prior mass, and the step embedding. The update is formulated as:
\begin{equation}
\begin{split}
    \mathbf{y}_{t+1}^{(i)} = & \, \mathbf{y}_t^{(i)}
    + \eta_{t,i} \sum_{j \in \mathcal{N}(i)} \alpha_{t,ij} (\mathbf{y}_t^{(j)} - \mathbf{y}_t^{(i)}) \\
    & + \beta_{t,i}(\mathbf{y}_{\text{init}}^{(i)}-\mathbf{y}_t^{(i)}) + \rho(\mathbf{h}_t^{(i)}),
\end{split}
\end{equation}
where $\alpha_{t,ij}$ denotes learned neighborhood attention weights, while $\beta_{t,i}$ and $\eta_{t,i}$ are adaptive bounded nonnegative pointwise coefficients regulating anchoring and diffusivity, respectively. The term $\rho(\mathbf{h}_t^{(i)})$ represents a residual refinement term predicted from the current node state. For the Graph Diffusion Refiner, $\mathcal{N}(i)$ is defined on a dynamically constructed fine-scale Euclidean $k$-NN graph with $k_{\mathrm{graph}}=24$, built once per forward pass and reused across the fixed $T=4$ diffusion steps. Moreover, $\alpha_{t,ij}$ is computed as normalized neighborhood attention weights from pairwise feature affinity with relative geometric bias, while the point-wise coefficients and residual term are predicted from the current node state. Thus, the module is a feature-guided anisotropic graph diffusion process rather than a fixed smoothing operator. All structural biases and diffusion coefficients are predicted by lightweight MLP heads from the corresponding point features, structural mask statistics, and global context.
After these fixed diffusion steps, we take the final state as the intensity logits, i.e., $\mathbf{y}_{\text{value}} \triangleq \mathbf{y}_T$. Subsequently, the final affordance map $\hat{\mathbf{R}}$ is synthesized via $\hat{\mathbf{R}} = \sigma(\mathbf{y}_{\text{value}}) \odot \sigma(\mathbf{y}_{\text{mask}})$. The entire framework is optimized end-to-end using the objective:
\begin{equation}
    \label{eq:5}
    \mathcal{L}_{\text{total}} = \mathcal{L}_{\text{main}} + \lambda_{m} \mathcal{L}_{\text{mask}} + \lambda_{v} \mathcal{L}_{\text{value}} + \lambda_{r} \mathcal{L}_{\text{rel}}.
\end{equation}
We aggregate reliability losses across scales as $\mathcal{L}_{\text{rel}} = \sum_{s \in \{c,m,f\}} \omega_s \mathcal{L}_{\text{rel}}^{s}$, where $\omega_s$ denotes the scale-specific reliability weight.
Here, $\mathcal{L}_{\text{main}}$ directly supervises the fused output using binary cross-entropy and MSE on the normalized affordance probabilities $\hat{\mathbf{R}}$. The auxiliary mask loss $\mathcal{L}_{\text{mask}}$ applies BCE supervision to $\mathbf{y}_{\text{mask}}$ with the binary target $\mathbb{I}[\mathbf{R}_{gt}\ge\delta_{\mathrm{valid}}]$. Crucially, the value loss is computed exclusively within valid regions where $\mathbf{R}_{gt} \ge \delta_{\mathrm{valid}}$:
\begin{equation}
\mathcal{L}_{\text{value}} =
\frac{1}{|\mathcal{V}|}\sum_{i \in \mathcal{V}}
\big|\sigma(\mathbf{y}_{\text{value}}^{(i)}) - \mathbf{R}_{gt}^{(i)}\big|,
\end{equation}
where $\mathcal{V} = \{i \mid \mathbf{R}_{gt}^{(i)} \ge \delta_{\mathrm{valid}}\}$ denotes the valid point set used for this loss.

\section{Dataset}
\begin{figure*}[t]
\centering
\begin{subfigure}[b]{\textwidth}
\includegraphics[width=\linewidth]{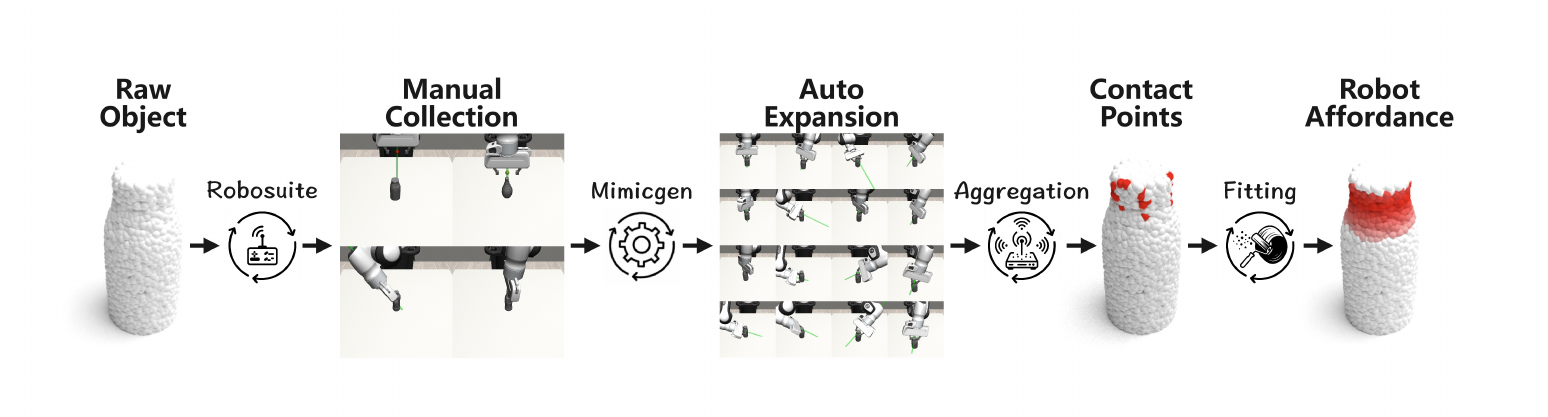}
\caption{Robot Affordance Generation Pipeline.}
\label{fig:hpad_build_pipeline}
\end{subfigure}\\[6pt]
\begin{subfigure}[b]{0.485\textwidth}
\includegraphics[width=\linewidth]{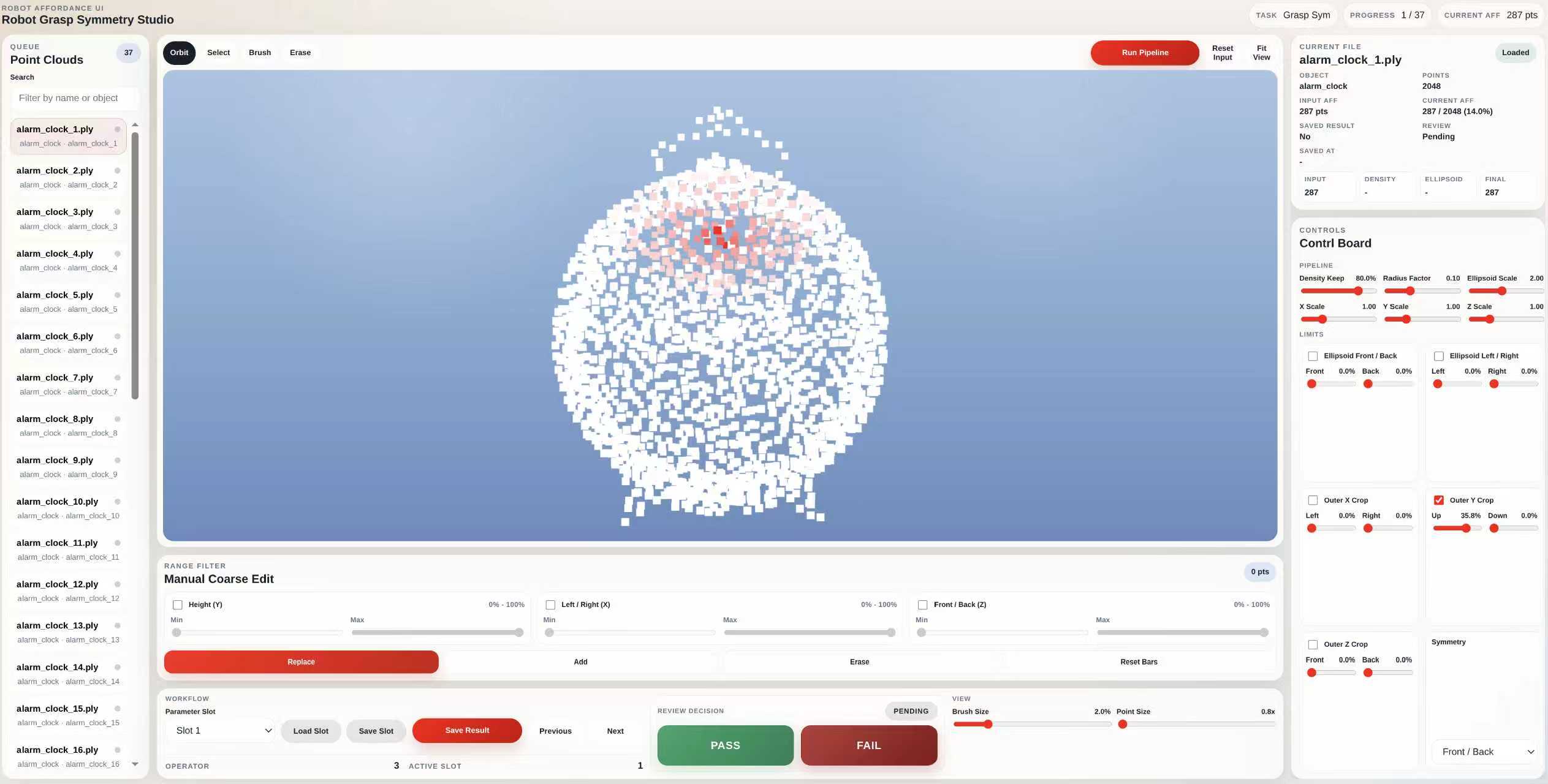}
\caption{Robot Affordance Refinement UI.}
\label{fig:hpad_build_robot}
\end{subfigure}\hfill
\begin{subfigure}[b]{0.485\textwidth}
\includegraphics[width=\linewidth]{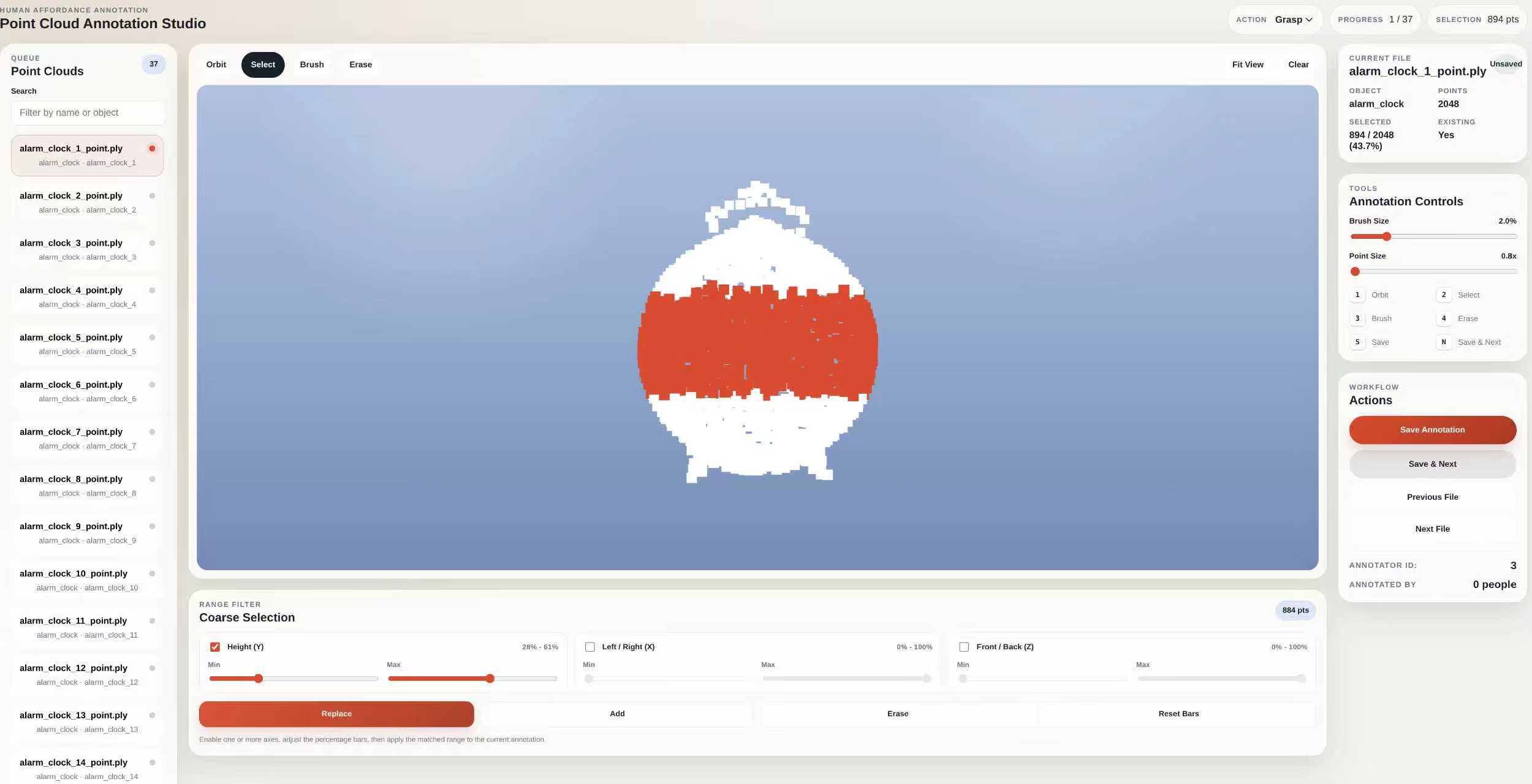}
\caption{Human Prior Annotation UI.}
\label{fig:hpad_build_human}
\end{subfigure}
\caption{\textbf{HPAD construction.} \textbf{(a)} Robot Affordance Generation Pipeline: an automated workflow that produces robot affordance labels from object point clouds. \textbf{(b)} Robot Affordance Refinement UI: an interactive tool used to refine the labels flagged by the quality checks. \textbf{(c)} Human Prior Annotation UI: a custom interface used to paint per-point human priors on the same object point clouds.}
\label{fig:hpad_build}
\end{figure*}

\begin{figure*}[t]
\centering
\begin{minipage}[t]{0.478\textwidth}
\vspace{0pt}%
\begin{subfigure}[t]{\linewidth}
\centering
\includegraphics[width=\linewidth]{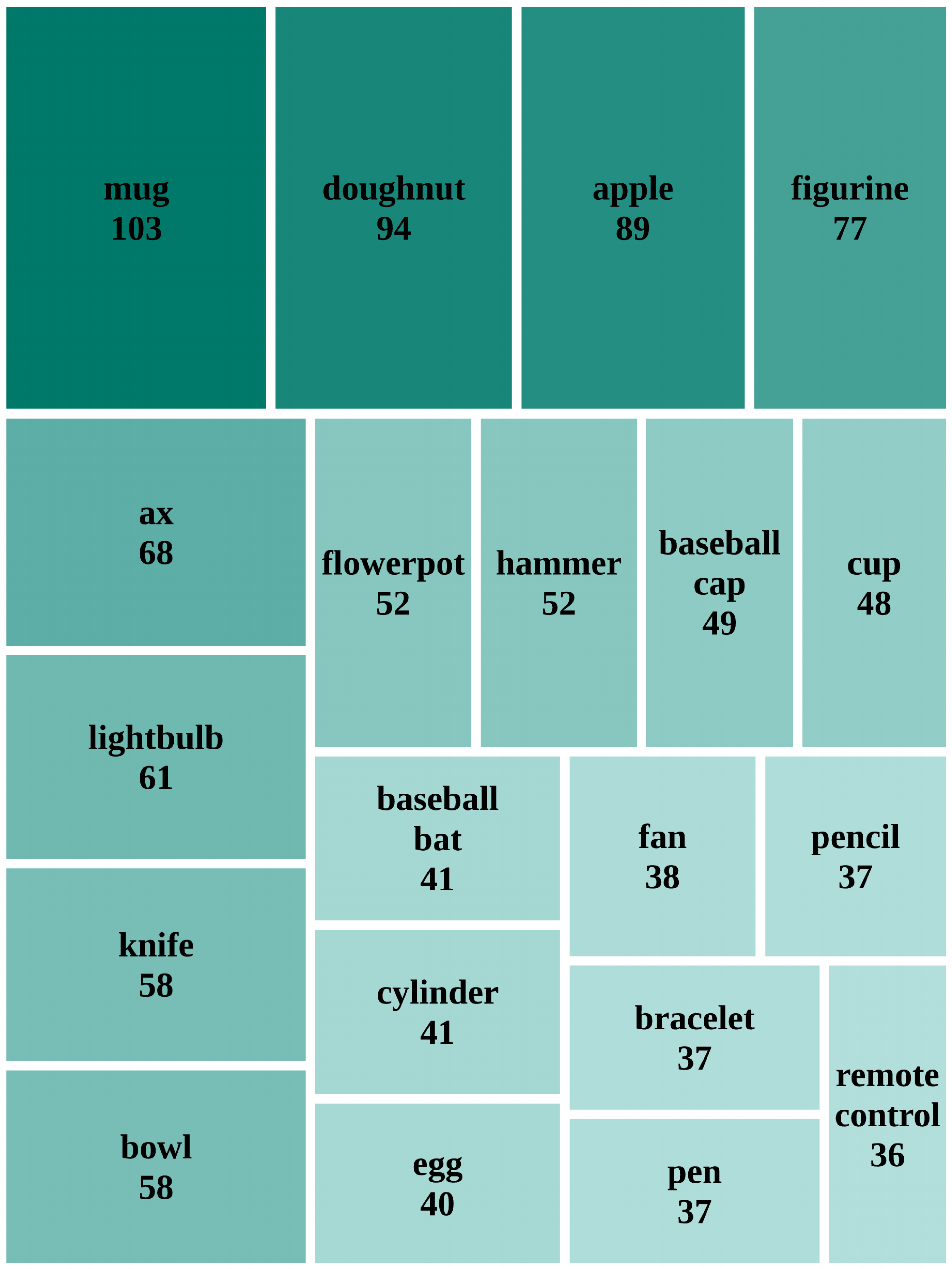}
\caption{Top-20 object categories.}
\label{fig:hpad_stats_categories}
\end{subfigure}
\end{minipage}\hspace{0.02\textwidth}%
\begin{minipage}[t]{0.481\textwidth}
\vspace{0pt}%
\begin{subfigure}[t]{\linewidth}
\centering
\includegraphics[width=0.98\linewidth,trim=33pt 40pt 33pt 41pt,clip]{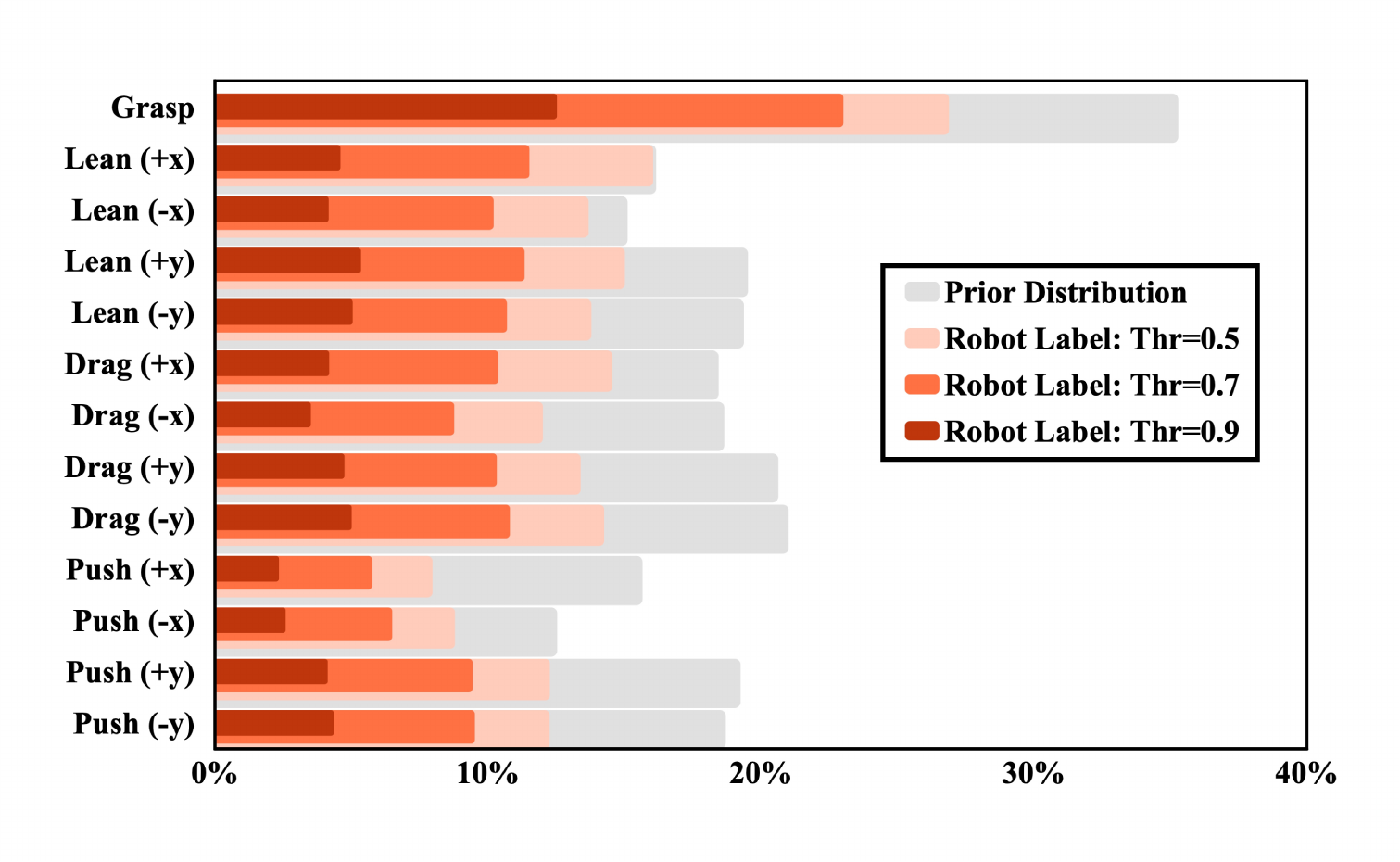}
\caption{Activation ratios.}
\label{fig:hpad_stats_activation}
\end{subfigure}

\vspace{0.62cm}

\begin{subfigure}[t]{\linewidth}
\centering
\includegraphics[width=0.85\linewidth,trim=35pt 40pt 39pt 49pt,clip]{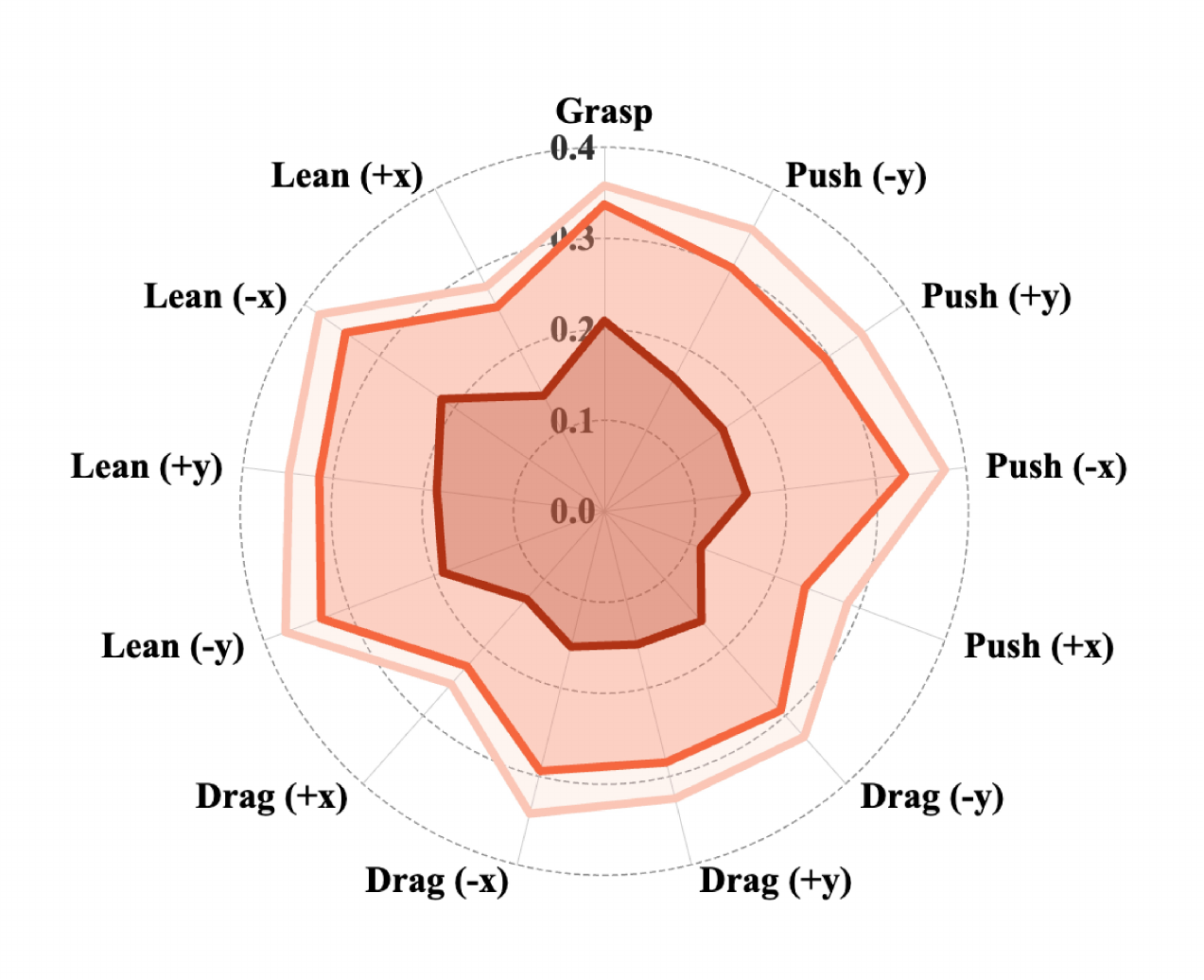}
\caption{IoU vs.\ threshold $\theta$.}
\label{fig:hpad_stats_iou}
\end{subfigure}
\end{minipage}
\caption{\textbf{HPAD statistics and the embodiment gap.} (a) Top-20 object categories by frequency. (b) Activation ratio of the binary human prior versus thresholded robot labels $\mathbb{I}[\mathbf{R}_{gt}\!\ge\!\theta]$ across $\theta$. (c) IoU between the same two activations as a function of $\theta$.}
\label{fig:hpad_stats}
\end{figure*}

\subsection{Dataset Definition}

\subsubsection{\texorpdfstring{\textbf{Object Curation}}{Object Curation}}
The object assets in HPAD are sourced from Objaverse \cite{objaverse}, a large-scale repository of 3D models. To ensure physical plausibility and relevance to tabletop manipulation, we filter the collection to a curated subset of 2,137 instances spanning 80 categories. All models are standardized via a unified preprocessing pipeline, including canonical pose alignment and physical-parameter specification, which improves simulation stability and enforces consistent orientation conventions across instances. Density and friction are assigned per category from the material of the real object and held fixed within it. The values are given in Appendix~B.

\subsubsection{\texorpdfstring{\textbf{Atomic Action Space}}{Atomic Action Space}}

\begin{table}[H]
\centering
\footnotesize
\renewcommand{\arraystretch}{1.18}
\setlength{\tabcolsep}{5pt}
\caption{\textbf{Specification of the atomic action primitives in HPAD.} Push, Drag, and Lean are reported over the four cardinal directions.}
\label{tab:atomic_action_summary}
\begin{tabularx}{\columnwidth}{l c >{\raggedright\arraybackslash}X}
\toprule
\rowcolor{mygray}
\textbf{Primitive} & \textbf{Dir.} & \textbf{Success criterion} \\
\midrule
\textbf{Grasp} & \textemdash             & Stable prehensile grasp; lift the object by \textbf{$\geq 5$\,cm} and hold for \textbf{$\geq 3$\,s}. \\
\cmidrule(lr){1-3}
\textbf{Push}  & $\pm x, \pm y$ & Approach from the opposite side; push the object by \textbf{$\geq 10$\,cm} while keeping tabletop contact. \\
\textbf{Drag}  & $\pm x, \pm y$ & Approach from the same side; drag the object by \textbf{$\geq 10$\,cm} while keeping tabletop contact. \\
\textbf{Lean}  & $\pm x, \pm y$ & Approach from the opposite side; tilt the object by \textbf{$\geq 15^{\circ}$} and keep the tilted pose stable for \textbf{$\geq 3$\,s} with partial bottom contact. \\
\bottomrule
\end{tabularx}
\end{table}

High-level functional semantics (e.g., ``drink'', ``sit'') convey abstract intent but omit the kinematic specificity required for direct robotic execution. To bridge this granularity gap, we define an action space of \textit{Atomic Action Primitives}. Concretely, we discretize physical interactions into 13 primitives: a pose-agnostic \textit{Grasp} and three directional operations (\textit{Push}, \textit{Drag}, \textit{Lean}) parameterized along the four cardinal axes of the canonical tabletop frame ($+x, -x, +y, -y$). We choose this set because it captures the dominant contact modes in tabletop manipulation, including prehensile grasping and non-prehensile interactions that impose contact constraints, while remaining simple enough to annotate and simulate at scale. We further discretize directional primitives into four axes to remove sign ambiguity in planar motion and to provide uniform, omni-directional coverage of the object surface under a fixed canonical frame. Tab.~\ref{tab:atomic_action_summary} lists the directions and quantitative success criteria for each primitive.

\subsection{Annotation Pipeline}
We first build the physically executable robot ground truth via a semi-automated pipeline, and then annotate the corresponding human prior on the same set of objects.

\subsubsection{\texorpdfstring{\textbf{Robot Affordance Annotation}}{Robot Affordance Annotation}}

To obtain physically executable ground truth, we construct a semi-automated annotation workflow based on RoboSuite~\cite{robosuite} and MimicGen~\cite{mimicgen}. As shown in Fig.~\ref{fig:hpad_build}(a), the Robot Affordance Generation Pipeline turns object point clouds into robot affordance labels through the stages described below; the low-quality cases are refined manually, as described at the end of this subsection.

\textit{\textbf{Demonstration sampling and expansion.}} For each (category, action) pair, we collect 5--10 short teleoperated seed demonstrations and leverage MimicGen to expand them into a large pool of candidate trajectories, yielding dense surface coverage.

\textit{\textbf{Task success verification.}} We enforce strict task-completion criteria (Tab.~\ref{tab:atomic_action_summary}) to filter the expanded trajectories. Across the 13 atomic primitives, on average about 30\% of expanded trials satisfy the criteria and are retained for supervision.

\textit{\textbf{Geometry-adaptive affordance fitting.}} We aggregate the discrete contact points from successful trials into continuous affordance heatmaps via density-based outlier rejection, anisotropic Gaussian smoothing, and symmetry-aware regularization. The automatic fitting already produces high-quality heatmaps for over 90\% of object--action pairs.

\textit{\textbf{Quality checks and manual refinement.}} Two checks are automatic: a pair enters heatmap fitting only after 20 successful trials (20 consecutive failures trigger re-collection), and a fitted heatmap is flagged when its outliers exceed 10\% of its contact points. Every heatmap is also reviewed manually. Flagged pairs, roughly 10\% of the total, are refined in the Robot Affordance Refinement UI (Fig.~\ref{fig:hpad_build}(b)), which loads the automatic fit, under three rules: start from that fit, derive the required symmetry from the object's geometry, and edit only locally without redrawing the region (full protocol in Appendix~A).

\subsubsection{\texorpdfstring{\textbf{Human Prior Annotation}}{Human Prior Annotation}}
For the source domain, following common practice in affordance benchmarks \cite{AffordanceNet21,IAGNet23,LASO24,nguyen2017,luo2022}, we obtain the human prior by annotation, as a per-point binary mask on the object point cloud \cite{AffordanceNet21} that marks where a person would interact with the object for a given action. This lightweight formulation avoids subjective fine-grained intensity scoring and reduces inter-annotator variance, so the human prior provides only coarse, macroscopic guidance rather than precise contact distributions. Five annotators, trained with a written guideline and a calibration batch, divided the 80 categories among themselves, each annotating entire categories for all 13 primitives. On the interface (Fig.~\ref{fig:hpad_build}(c)), which supports point-cloud rotation, region painting, action-conditioned panel switching, and per-action review, they marked a rough region and then refined it with a brush. A second annotator verified every mask. Disputed masks were corrected, and residual disagreements were settled by the project leader (full protocol in Appendix~A).

\subsection{Statistical Analysis}

HPAD contains over 27{,}000 object--action pairs spanning a wide range of categories; Fig.~\ref{fig:hpad_stats}(a) summarizes the top-20 categories by frequency.
Our analysis highlights systematic embodiment gaps between human priors and robot-executable labels.
As shown in Fig.~\ref{fig:hpad_stats}(b), human priors generally activate larger regions than robot labels, reflecting the broader tolerance of human hand interactions.
Moreover, Fig.~\ref{fig:hpad_stats}(c) reveals substantial spatial misalignment: the IoU drops from above 0.3 to below 0.2 as the confidence threshold increases.
Notably, this discrepancy persists even in high-confidence regions, suggesting that closing the ``semantic--physical'' gap requires an explicit transfer mechanism rather than a direct mapping. The sensitivity of the robot labels to the simulation parameters is analyzed in Appendix~C.
\begin{table*}[!t]
\centering
\caption{\textbf{Comparison with existing 3D affordance datasets.} \datasettri{} indicates part-level or action-related annotations that can support manipulation heuristics, rather than dense robot affordance maps.}
\label{tab:dataset_comparison}
\resizebox{\textwidth}{!}{%
\begin{tabular}{lccccccc}
\toprule
\textbf{Dataset} & \textbf{\#Obj.} & \textbf{\#Cat.} & \textbf{\#Act.} & \textbf{Human Prior} & \textbf{Robot Labels} & \textbf{Paired H-R} & \textbf{Robot Validation} \\
\midrule
3D AffordanceNet~\cite{AffordanceNet21} & 22{,}949 & 23 & 18 & \datasetcmark & \datasetxmark & \datasetxmark & \datasetxmark \\
PIAD~\cite{IAGNet23} & 7{,}012 & 23 & 17 & \datasetcmark & \datasetxmark & \datasetxmark & \datasetxmark \\
AffordPose~\cite{jian2023affordpose} & 641 & 13 & 8 & \datasetcmark & \datasetxmark & \datasetxmark & \datasetxmark \\
LASO~\cite{LASO24} & 8{,}434 & 23 & 17 & \datasetcmark & \datasetxmark & \datasetxmark & \datasetxmark \\
Where2Act~\cite{where2act} & 972 & 15 & 6 & \datasetxmark & \datasetcmark & \datasetxmark & \datasetcmark \\
GAPartNet~\cite{geng2023gapartnet} & 1{,}166 & 27 & 9 & \datasetxmark & \datasettri & \datasetxmark & \datasetxmark \\
\textbf{HPAD (Ours)} & \textbf{2{,}137} & \textbf{80} & \textbf{13} & \textbf{\datasetcmark} & \textbf{\datasetcmark} & \textbf{\datasetcmark} & \textbf{\datasetcmark} \\
\bottomrule
\end{tabular}%
}
\end{table*}

For comparison with existing affordance datasets, Tab.~\ref{tab:dataset_comparison} positions HPAD against representative 3D affordance datasets. Compared with prior datasets, HPAD is the only one in this comparison that simultaneously provides paired human priors, robot labels, and robot validation at the object level, while also covering substantially more object categories and defining a robot-executable atomic action space. This design makes HPAD a testbed for studying how human interaction knowledge can be transformed into robot-executable affordance distributions, enabling controlled evaluation of both cross-embodiment transfer and downstream actionability.

To comprehensively probe model performance beyond in-distribution accuracy and to assess generalization, we split HPAD into \textbf{Seen} and \textbf{Unseen} partitions. The \textbf{Seen} partition (1{,}787 train / 350 val objects) shares its object categories between training and validation, evaluating intra-category generalization to unseen instances of categories that have already been observed during training. The \textbf{Unseen} partition (1{,}500 train / 250 val objects spanning 55 / 21 categories) holds out entire object categories at validation time, evaluating cross-category generalization to object categories that were never observed during training.

\section{Experiment}

\subsection{Implementation Details}
We implement ART in PyTorch. The input point cloud $\mathbf{P}$ consists of $N=2048$ points normalized to the unit sphere. The human prior $\mathbf{H}$ is appended as an additional input channel, while the object size descriptor is set to $\mathbf{d}=\log(\mathbf{b}+\epsilon)$, where $\mathbf{b}$ is the pre-normalization axis-aligned bounding-box dimension and $\epsilon=10^{-6}$. The encoded descriptor is used to condition the global context $\mathbf{g}$. The backbone is PointNeXt-S~\cite{pointnext}, with a stem width of 64, four stages with widths $[64,128,256,512]$, and block depths $[2,2,3,1]$. The learnable action-indexed robot prototypes $\mathbf{Z}_a$ contain 16 tokens per action, each with dimension 128, and are linearly projected to the attention dimension at each HAR scale. The Graph Diffusion Refiner uses $T=4$ anisotropic diffusion steps on the dynamically constructed $k$-NN graph with $k_{\mathrm{graph}}=24$ and 4 attention heads. We instantiate the multiscale hierarchy $\{c,m,f\}$ at point resolutions of $64$, $256$, and $2048$, respectively.

We train ART in an action-conditioned joint setting, where samples from all 13 atomic action primitives are optimized together and the action condition $a$ modulates the global context and selects the corresponding robot-prototype bank. This is a single joint model rather than a set of separately trained action models. The model is trained for 200 epochs using AdamW with a batch size of 44 per GPU, giving an effective batch size of 352 across 8 NVIDIA RTX 4090 GPUs. We use a cosine annealing schedule decaying to $10^{-6}$ after a 5-epoch linear warmup. The learning rate is set to $10^{-3}$, weight decay to $10^{-2}$, and gradients are clipped at an $\ell_2$ norm of 1.0. Under this configuration, training the joint action-conditioned model takes approximately 4 hours on the 8-GPU node.

The total objective in Eq.~\ref{eq:5} uses $\lambda_m=0.2$, $\lambda_v=0.1$, and $\lambda_r=0.1$. For the main supervision $\mathcal{L}_{\text{main}}$, BCE and MSE terms are combined with relative weights of 0.7 and 0.3, respectively. We set $\delta_{\mathrm{valid}}=0.1$ for the mask and positive-region value losses. For the reliability term, robot-affordance targets of at least 0.01 are treated as positives in the weighted BCE, yielding a class-balancing ratio of 1:3 (pos:neg) and scale weights $(\omega_c,\omega_m,\omega_f)=(0.5,0.5,1.0)$ for the $64/256/2048$ point resolutions.

\subsection{Benchmark Setting}
\subsubsection{\texorpdfstring{\textbf{Evaluation Metrics}}{Evaluation Metrics}}
Following established protocols~\cite{IAGNet23,AffordanceNet21}, we report \textbf{MAE}, \textbf{aIoU}, $\mathbf{IoU}_{0.9}$, and $\mathbf{F1}_{0.9}$, defined below. Throughout, $\hat{\mathbf{R}}\in[0,1]^N$ denotes the predicted per-point robot affordance and $\mathbf{R}_{gt}\in[0,1]^N$ the simulation-validated continuous ground truth. In all quantitative tables, aIoU, IoU$_{0.9}$, and F1$_{0.9}$ are reported as percentages, while MAE is reported on its original $[0,1]$ scale.

\textit{\textbf{Mean Absolute Error (MAE)}}
\begin{equation}
\mathrm{MAE}=\frac{1}{N}\sum_{i=1}^{N}\bigl|\hat{R}_i-R_{gt,i}\bigr|.
\end{equation}
MAE is the per-point absolute deviation between the predicted heatmap $\hat{\mathbf{R}}$ and the continuous ground truth $\mathbf{R}_{gt}$, averaged over all $N$ points, capturing how faithfully the predicted distribution tracks the simulation-validated robot affordance across the entire object surface before any thresholding.

\textit{\textbf{Average Intersection over Union (aIoU)}}
\begin{equation}
\mathrm{aIoU}=\frac{1}{|\Theta|}\sum_{\theta\in\Theta}\frac{|\mathcal{S}_{\mathrm{pred}}(\theta)\cap\mathcal{S}_{\mathrm{gt}}(\theta)|}{|\mathcal{S}_{\mathrm{pred}}(\theta)\cup\mathcal{S}_{\mathrm{gt}}(\theta)|},
\end{equation}
where $\mathcal{S}_{\mathrm{pred}}(\theta)=\{i:\hat{R}_i>\theta\}$ and $\mathcal{S}_{\mathrm{gt}}(\theta)=\{i:R_{gt,i}>\theta\}$ are the predicted and ground-truth high-confidence point sets at threshold $\theta$, and $\Theta=\{0.10, 0.11, \ldots, 0.99\}$ ($|\Theta|=90$). aIoU averages, over $\Theta$, the IoU between $\mathcal{S}_{\mathrm{pred}}(\theta)$ and $\mathcal{S}_{\mathrm{gt}}(\theta)$, measuring threshold-agnostic agreement at matched confidence levels across the full operating range.

\textit{\textbf{High-Confidence IoU} ($\mathbf{IoU}_{0.9}$)}
\begin{equation}
\mathrm{IoU}_{0.9}=\frac{|\mathcal{S}_{\mathrm{pred}}(0.9)\cap\mathcal{S}_{\mathrm{gt}}(0.9)|}{|\mathcal{S}_{\mathrm{pred}}(0.9)\cup\mathcal{S}_{\mathrm{gt}}(0.9)|},
\end{equation}
where $\mathcal{S}_{\mathrm{pred}}(0.9)$ and $\mathcal{S}_{\mathrm{gt}}(0.9)$ are the predicted and ground-truth high-confidence point sets at threshold $\theta=0.9$, instantiating the aIoU family at its most stringent operating point. We focus on the $\theta=0.9$ regime because at execution time the robot preferentially samples contact candidates from the most confident regions of the predicted affordance. Agreement at this regime is therefore what determines whether the predicted distribution is actionable, complementing aIoU's threshold-agnostic view.

\textit{\textbf{High-Confidence F1} ($\mathbf{F1}_{0.9}$)}
\begin{equation}
\mathrm{F1}_{0.9}=\frac{2\,|\mathcal{S}_{\mathrm{pred}}(0.9)\cap\mathcal{S}_{\mathrm{gt}}(0.9)|}{|\mathcal{S}_{\mathrm{pred}}(0.9)|+|\mathcal{S}_{\mathrm{gt}}(0.9)|},
\end{equation}
where $\mathcal{S}_{\mathrm{pred}}(0.9)$ and $\mathcal{S}_{\mathrm{gt}}(0.9)$ are the same high-confidence point sets used by $\mathrm{IoU}_{0.9}$. $\mathrm{F1}_{0.9}$ is reported alongside $\mathrm{IoU}_{0.9}$ as a complementary measure: as the harmonic mean of precision and recall on the high-confidence masks, it is less sensitive to size mismatch between $\mathcal{S}_{\mathrm{pred}}(0.9)$ and $\mathcal{S}_{\mathrm{gt}}(0.9)$ and exposes whether the agreement is balanced rather than dominated by over- or under-prediction.

\subsubsection{\texorpdfstring{\textbf{Modular Baselines}}{Modular Baselines}}
To the best of our knowledge, prior work has not directly leveraged human affordance priors to predict robot-executable 3D affordance distributions. We therefore construct a set of modular baselines built on widely used PointNet++~\cite{pointnet2} and PointNeXt~\cite{pointnext} backbones to benchmark three fusion strategies: \textbf{(1)} \textbf{\textcolor{cBlue}{Robot-Only}}, which takes only the object point cloud as input to assess the necessity of human guidance; \textbf{(2)} \textbf{\textcolor{cGreen}{Input Concatenation}}, which appends the per-point human prior map as an additional input channel before the backbone; and \textbf{(3)} \textbf{\textcolor{cOrange}{Feature Fusion}}, which injects the prior at the latent level on the stronger PointNeXt backbone via conditioning modules, including Cross-Attention~\cite{attention} and FiLM~\cite{film}. For a fair comparison, all baselines additionally take the action index and object size as inputs; we encode them, broadcast them to all points, and concatenate them to the per-point input features.
\begin{table*}[!t]
\centering
\caption{Comparison results on HPAD. The columns are color-coded by prior-integration strategy: \textbf{\textcolor{cBlue}{Blue}} denotes methods that do not use the human prior (\textbf{\textcolor{cBlue}{Robot-Only}}); \textbf{\textcolor{cGreen}{Green}} denotes \textbf{\textcolor{cGreen}{Input Concatenation}}, where the per-point human prior map is appended as an input channel; and \textbf{\textcolor{cOrange}{Orange}} denotes \textbf{\textcolor{cOrange}{Feature Fusion}} baselines that inject the prior at the latent level via conditioning modules (Cross-Attention/FiLM) on the PointNeXt backbone. $\textcolor{darkpink}{\diamond}$ denotes the relative improvement of our method. \textbf{Bold} marks the best result. \textbf{Ours w/o prior} is ART trained and evaluated with the human prior set to zero.}

\resizebox{\textwidth}{!}{
\setlength{\tabcolsep}{4pt}
\begin{tabular}{c|l|cc|cc|cc|c|c}
\toprule
 & \textbf{Metrics}
 & \textbf{\textcolor{cBlue}{PointNet++}}
 & \textbf{\textcolor{cBlue}{PointNeXt}}
 & \textbf{\textcolor{cGreen}{PointNet++}}
 & \textbf{\textcolor{cGreen}{PointNeXt}}
 & \textbf{\textcolor{cOrange}{Cross}}
 & \textbf{\textcolor{cOrange}{FiLM}}
 & \textbf{\textcolor{cBlue}{Ours w/o prior}}
 & \textbf{Ours} \\
\midrule

\multirow{4}{*}{\rotatebox{90}{\textbf{\texttt{Seen}}}}
    & MAE $\downarrow$
    & $0.041\textcolor{darkpink}{\diamond22.0\%}$
    & $0.038\textcolor{darkpink}{\diamond15.8\%}$
    & $0.040\textcolor{darkpink}{\diamond20.0\%}$
    & $0.037\textcolor{darkpink}{\diamond13.5\%}$
    & $0.036\textcolor{darkpink}{\diamond11.1\%}$
    & $0.036\textcolor{darkpink}{\diamond11.1\%}$
    & $0.034\textcolor{darkpink}{\diamond5.9\%}$
    & \cellcolor{mygray}\textbf{0.032} \\

    & aIoU (\%) $\uparrow$
    & $60.65\textcolor{darkpink}{\diamond13.9\%}$
    & $63.69\textcolor{darkpink}{\diamond8.4\%}$
    & $60.80\textcolor{darkpink}{\diamond13.6\%}$
    & $64.10\textcolor{darkpink}{\diamond7.7\%}$
    & $64.56\textcolor{darkpink}{\diamond7.0\%}$
    & $64.95\textcolor{darkpink}{\diamond6.3\%}$
    & $66.95\textcolor{darkpink}{\diamond3.2\%}$
    & \cellcolor{mygray}\textbf{69.06} \\

    & IoU$_{0.9}$ (\%) $\uparrow$
    & $38.08\textcolor{darkpink}{\diamond48.2\%}$
    & $44.57\textcolor{darkpink}{\diamond26.7\%}$
    & $37.41\textcolor{darkpink}{\diamond50.9\%}$
    & $44.69\textcolor{darkpink}{\diamond26.3\%}$
    & $46.11\textcolor{darkpink}{\diamond22.4\%}$
    & $46.56\textcolor{darkpink}{\diamond21.2\%}$
    & $51.07\textcolor{darkpink}{\diamond10.5\%}$
    & \cellcolor{mygray}\textbf{56.45} \\

    & F1$_{0.9}$ (\%) $\uparrow$
    & $47.92\textcolor{darkpink}{\diamond41.7\%}$
    & $55.38\textcolor{darkpink}{\diamond22.6\%}$
    & $47.26\textcolor{darkpink}{\diamond43.6\%}$
    & $55.74\textcolor{darkpink}{\diamond21.8\%}$
    & $57.16\textcolor{darkpink}{\diamond18.8\%}$
    & $57.83\textcolor{darkpink}{\diamond17.4\%}$
    & $62.48\textcolor{darkpink}{\diamond8.6\%}$
    & \cellcolor{mygray}\textbf{67.88} \\
\midrule

\multirow{4}{*}{\rotatebox{90}{\textbf{\texttt{Unseen}}}}
    & MAE $\downarrow$
    & $0.139\textcolor{darkpink}{\diamond27.3\%}$
    & $0.130\textcolor{darkpink}{\diamond22.3\%}$
    & $0.113\textcolor{darkpink}{\diamond10.6\%}$
    & $0.113\textcolor{darkpink}{\diamond10.6\%}$
    & $0.111\textcolor{darkpink}{\diamond9.0\%}$
    & $0.111\textcolor{darkpink}{\diamond9.0\%}$
    & $0.129\textcolor{darkpink}{\diamond21.7\%}$
    & \cellcolor{mygray}\textbf{0.101} \\

    & aIoU (\%) $\uparrow$
    & $27.54\textcolor{darkpink}{\diamond59.4\%}$
    & $33.79\textcolor{darkpink}{\diamond29.9\%}$
    & $34.75\textcolor{darkpink}{\diamond26.3\%}$
    & $37.33\textcolor{darkpink}{\diamond17.6\%}$
    & $38.30\textcolor{darkpink}{\diamond14.6\%}$
    & $38.79\textcolor{darkpink}{\diamond13.2\%}$
    & $34.44\textcolor{darkpink}{\diamond27.5\%}$
    & \cellcolor{mygray}\textbf{43.90} \\

    & IoU$_{0.9}$ (\%) $\uparrow$
    & $11.97\textcolor{darkpink}{\diamond148.1\%}$
    & $14.66\textcolor{darkpink}{\diamond102.6\%}$
    & $15.59\textcolor{darkpink}{\diamond90.5\%}$
    & $18.78\textcolor{darkpink}{\diamond58.1\%}$
    & $21.82\textcolor{darkpink}{\diamond36.1\%}$
    & $20.86\textcolor{darkpink}{\diamond42.4\%}$
    & $18.48\textcolor{darkpink}{\diamond60.7\%}$
    & \cellcolor{mygray}\textbf{29.70} \\

    & F1$_{0.9}$ (\%) $\uparrow$
    & $15.90\textcolor{darkpink}{\diamond145.8\%}$
    & $19.58\textcolor{darkpink}{\diamond99.6\%}$
    & $21.06\textcolor{darkpink}{\diamond85.6\%}$
    & $25.32\textcolor{darkpink}{\diamond54.3\%}$
    & $29.15\textcolor{darkpink}{\diamond34.1\%}$
    & $27.61\textcolor{darkpink}{\diamond41.5\%}$
    & $24.49\textcolor{darkpink}{\diamond59.6\%}$
    & \cellcolor{mygray}\textbf{39.08} \\
\bottomrule
\end{tabular}
} 
\label{table:results}
\end{table*}
\begin{figure*}[t]
    \centering
    \begin{overpic}[width=\textwidth]{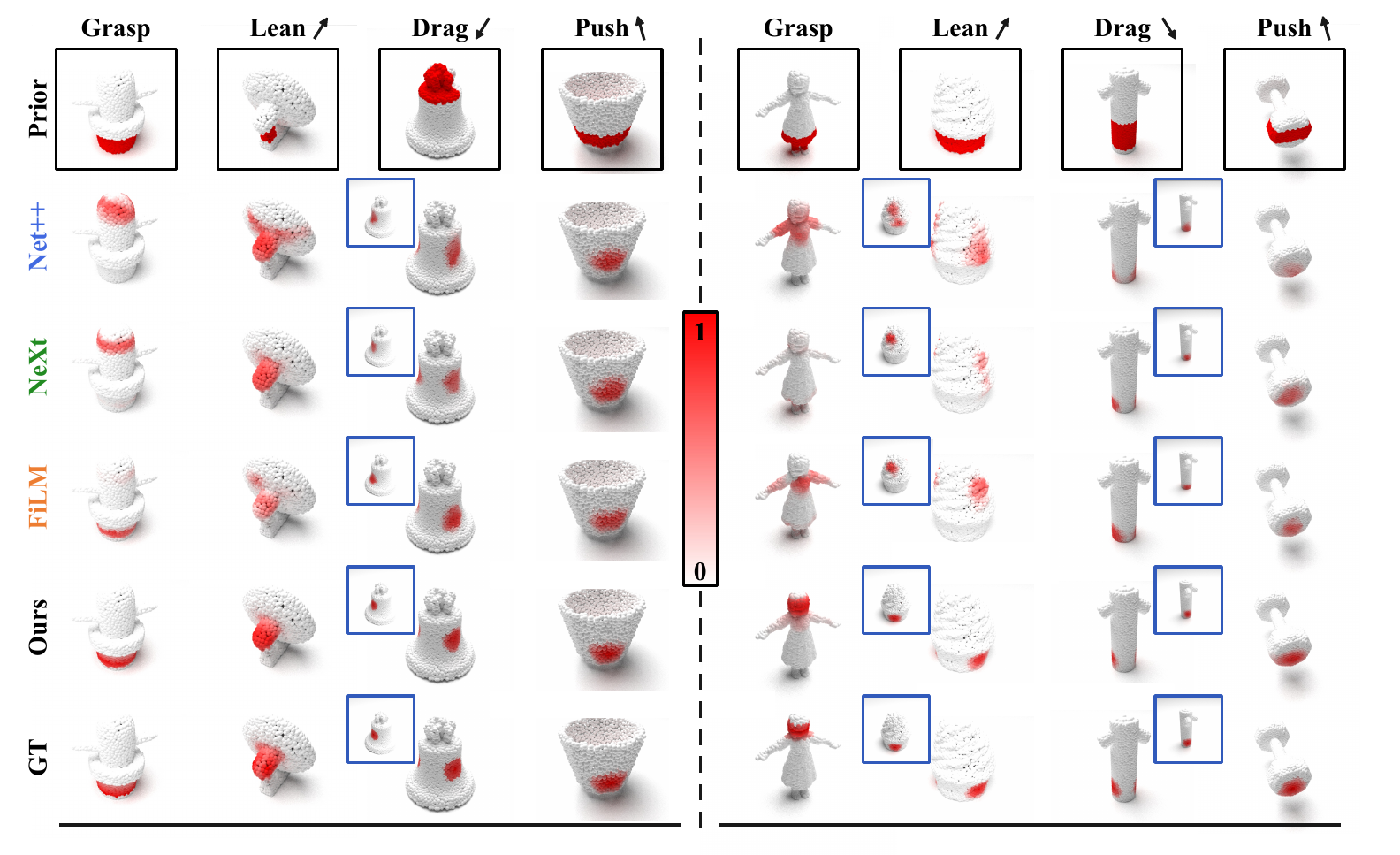}
        \put(25,0.7){\makebox(0,0){\textbf{\texttt{Seen}}}}
        \put(75,0.7){\makebox(0,0){\textbf{\texttt{Unseen}}}}
    \end{overpic}
    \caption{\textbf{Visualization results.} Rows show the human prior (\textbf{Prior}), four method-class predictions from \textbf{\textcolor{cBlue}{Net++}}, \textbf{\textcolor{cGreen}{NeXt}}, \textbf{\textcolor{cOrange}{FiLM}}, and \textbf{Ours}, and the ground truth (\textbf{GT}). Left four columns are drawn from the Seen partition and right four from the Unseen partition. \textbf{Arrows} $\bm{\uparrow}$ indicate the action direction, and red intensity encodes the per-point affordance on the central $[0,1]$ legend. Inset boxes show the back-side view of the same object, revealing affordance regions partly occluded in the main view.}
    \label{fig:viz_main}

\end{figure*}
\subsection{Comparison Results}
Tab.~\ref{table:results} reports the comparison results on HPAD. We discuss the Seen and Unseen partitions separately, since each isolates a distinct aspect of the proposed transfer setting.

\subsubsection{\texorpdfstring{\textbf{Quantitative Analysis on Seen Categories}}{Quantitative Analysis on Seen Categories}}
The Seen split fixes object categories between training and validation, the regime where weaker baselines should be hardest to distinguish from stronger ones. Yet on Seen ART already establishes a wide gap, and the shape of that gap is informative. Relative to FiLM, the strongest or tied-strongest baseline on every Seen metric, ART reduces MAE by $11.1\%$ and improves aIoU by $6.3\%$, IoU$_{0.9}$ by $21.2\%$, and F1$_{0.9}$ by $17.4\%$---the relative lead more than triples from threshold-averaged aIoU to high-confidence IoU$_{0.9}$. The widening is informative: aIoU averages agreement over the full threshold range while IoU$_{0.9}$ measures it at the most stringent operating point, so the gradient indicates that ART's advantage is most pronounced where the predictions are most committed. Cross-Attention and FiLM, both PointNeXt-based and purpose-built to fuse the human prior, reach aIoU close to ART (64.56 / 64.95 vs.\ 69.06) but stall on the high-confidence metrics (IoU$_{0.9}$ around 46, F1$_{0.9}$ around 57), reproducing the surface-like topology of the human prior even with full category-level evidence. ART reaches 56.45 and 67.88, indicating that the gap is structural rather than incremental: a transfer mechanism that resolves the human-to-robot topological mismatch, consistent with the design intent of GRG and HAR.

\subsubsection{\texorpdfstring{\textbf{Quantitative Analysis on Unseen Categories}}{Quantitative Analysis on Unseen Categories}}
The Unseen split holds out entire object categories at validation time, the regime where transfer must work without category-level shortcuts. Unseen does not merely raise the difficulty; it isolates the same mechanism Seen revealed, by removing the in-distribution evidence that lets weaker mechanisms coast. ART's lead expands accordingly. Relative to FiLM, ART reduces MAE by $9.0\%$ and improves aIoU by $13.2\%$, IoU$_{0.9}$ by $42.4\%$, and F1$_{0.9}$ by $41.5\%$; the high-confidence lead doubles relative to Seen ($21.2\%\!\to\!42.4\%$; $17.4\%\!\to\!41.5\%$). Cross-Attention edges FiLM on the high-confidence metrics, yet ART still leads Cross-Attention by $36.1\%$ / $34.1\%$, so the margin holds against every conditioning baseline. That the high-confidence margin grows precisely when in-distribution scaffolding is removed is direct evidence that the structural advantage seen on Seen is general-purpose, not category-memorized. ART simultaneously holds the highest absolute Unseen score on every metric and the smallest Seen-to-Unseen relative drop on aIoU and the high-confidence metrics (aIoU drops: ART $36.4\%$ vs.\ FiLM $40.3\%$; IoU$_{0.9}$ drops: ART $47.4\%$ vs.\ Cross-Attention $52.7\%$), ruling out a floor-effect explanation.

\subsubsection{\texorpdfstring{\textbf{Necessity of the Human Prior}}{Necessity of the Human Prior}}
The task setting we propose rests on a question Tab.~\ref{table:results} answers directly: does injecting a human prior actually improve robot-affordance inference? The table answers it along two axes: what the prior adds to the baselines, which inject it into a Robot-Only backbone by concatenation, Cross-Attention or FiLM, and what it adds to ART itself, which we ablate by setting the prior input to zero in training and evaluation (\textbf{Ours w/o prior}). On Seen, training instances of the same category already give the backbone a strong geometric basis, so the raw prior adds little to the baselines. ART nonetheless gains from it, mostly on the high-confidence metrics that govern execution. On Unseen, the prior becomes decisive: concatenation alone raises every baseline well above the Robot-Only backbone, ART widens this margin further, and ART without the prior falls back to roughly the level of the Robot-Only baseline in aIoU. At the task level, therefore, a human prior improves robot-affordance prediction by itself, most clearly on categories the model has never seen. At the model level, ART, whose mechanisms are designed for affordance transfer, extracts a larger gain from the same prior than any baseline on both splits, which confirms the effectiveness of its design.

\subsubsection{\texorpdfstring{\textbf{Qualitative Analysis}}{Qualitative Analysis}}
Fig.~\ref{fig:viz_main} corroborates Tab.~\ref{table:results} at the per-instance level and surfaces the same three-tier separation between method classes. Without a human prior, \textbf{\textcolor{cBlue}{PointNet++}} (Robot-Only) drifts away from actionable contact regions, often placing high-confidence mass on geometrically prominent but action-irrelevant surfaces. Once the prior is supplied, \textbf{\textcolor{cGreen}{PointNeXt}} (Concat) and \textbf{\textcolor{cOrange}{FiLM}} land on the right region of the object but spread their mass along the surface-like topology of the prior itself, producing diffuse low-saturation responses that fail to concentrate on the local area the robot would actually contact. ART produces visibly sharper and more localized predictions whose high-confidence support tracks GT closely on both partitions, including in the back-side views shown by the inset boxes. This three-tier visual gradient over Robot-Only, generic prior fusion, and structural transfer matches the quantitative gradient on the high-confidence metrics.

\subsection{Ablation Study}

\begin{table}[h]
\centering
\caption{\textbf{Ablation Study.} We investigate the contribution of the \textbf{GRG} and \textbf{HAR} on the HPAD dataset.}
\label{table:ablation}
\resizebox{\linewidth}{!}{
\setlength{\tabcolsep}{3.5pt}
\renewcommand{\arraystretch}{1.18}
\begin{tabular}{c|cc|cccc}
\toprule
\multicolumn{1}{l|}{} & \textbf{GRG} & \textbf{HAR} & \textbf{MAE} $\downarrow$ & \textbf{aIoU (\%)} $\uparrow$ & \textbf{IoU$_{0.9}$ (\%)} $\uparrow$ & \textbf{F1$_{0.9}$ (\%)} $\uparrow$ \\
\midrule
\multirow{4}{*}{\rotatebox{90}{\textbf{\texttt{Seen}}}}
 &  &  & 0.036 & 65.63 & 47.78 & 58.97 \\ 
 & \checkmark &  & 0.034 & 66.09 & 48.52 & 59.89 \\ 
 &  & \checkmark & 0.034 & 66.99 & 51.50 & 62.86 \\ 
 & \checkmark & \checkmark & \cellcolor{mygray}\textbf{0.032} & \cellcolor{mygray}\textbf{69.06} & \cellcolor{mygray}\textbf{56.45} & \cellcolor{mygray}\textbf{67.88} \\ 
\midrule
\multirow{4}{*}{\rotatebox{90}{\textbf{\texttt{Unseen}}}}
 &  &  & 0.108 & 37.81 & 19.89 & 26.41 \\ 
 & \checkmark &  & 0.105 & 39.75 & 24.04 & 31.83 \\ 
 &  & \checkmark & 0.106 & 40.29 & 25.29 & 33.68 \\ 
 & \checkmark & \checkmark & \cellcolor{mygray}\textbf{0.101} & \cellcolor{mygray}\textbf{43.90} & \cellcolor{mygray}\textbf{29.70} & \cellcolor{mygray}\textbf{39.08} \\ 
\bottomrule
\end{tabular}
}
\end{table}

\subsubsection{\texorpdfstring{\textbf{Effectiveness of GRG}}{Effectiveness of GRG}}

\begin{figure}[t]
    \centering
    \begin{overpic}[width=\columnwidth]{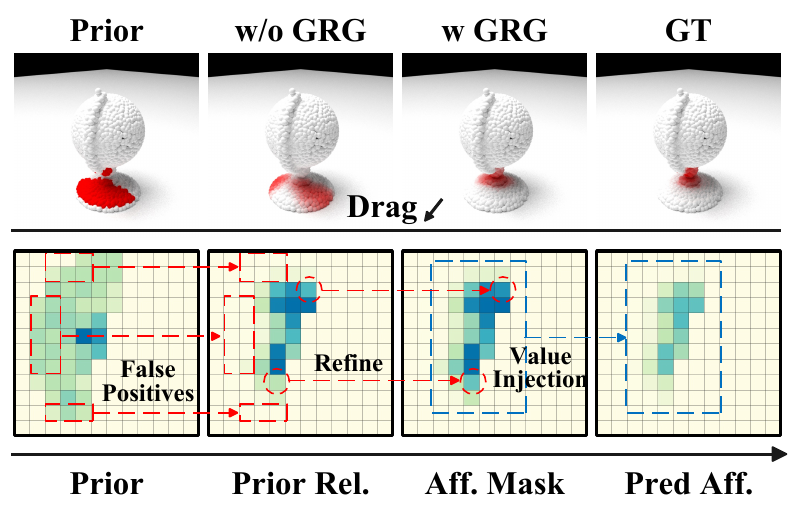}

    \end{overpic}
    \vspace{-3mm}
    \caption{\textbf{Ablation of GRG.} The top row shows qualitative results of the transferred affordance with and without GRG. The bottom row visualizes the intermediate feature evolution induced by GRG.}
    \vspace{-5mm}
    \label{fig:ab1}
\end{figure}

Results in Tab.~\ref{table:ablation} and Fig.~\ref{fig:ab1} highlight the importance of GRG. Without GRG, the model cannot effectively suppress geometric false positives inherited from the coarse human prior, leading to over-activated, non-executable regions. Adding GRG alone improves IoU$_{0.9}$ by $+20.9\%$ and F1$_{0.9}$ by $+20.5\%$ on the Unseen split, where the human prior---shaped by the high degrees of freedom of the human hand---introduces systematic geometric biases when applied to rigid robot grippers. GRG's learned reliability gating directly targets these biases by synergizing the human prior with global geometric features to construct a physical validity critic. The improvement concentrates on high-confidence metrics (IoU$_{0.9}$, F1$_{0.9}$) rather than threshold-averaged aIoU, indicating that GRG sharpens the actionable boundary rather than merely adjusting the overall distribution. This pattern holds on the Seen split, confirming that the filtering mechanism generalizes across partitions. Fig.~\ref{fig:ab1} illustrates this mechanism: the top row compares predictions with and without GRG, showing that without GRG many false positives from the human prior remain unfiltered, while the bottom row visualizes how GRG progressively attenuates spurious activations while preserving geometrically plausible contacts.

\subsubsection{\texorpdfstring{\textbf{Effectiveness of HAR}}{Effectiveness of HAR}}

\begin{figure}[t]
    \centering
    \begin{overpic}[width=\columnwidth]{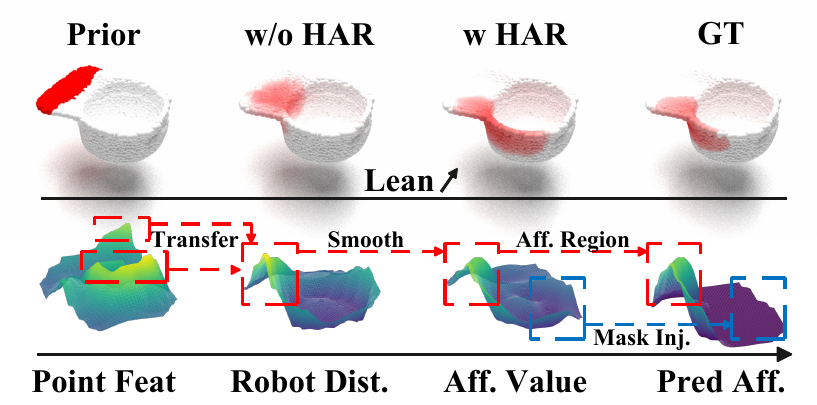}
    \end{overpic}
    \caption{\textbf{Ablation of HAR.} The top row shows qualitative results of the transferred affordance with and without HAR. The bottom row visualizes the intermediate feature evolution.}
    \label{fig:ab2}
\vspace{-4mm}

\end{figure}

Results in Tab.~\ref{table:ablation} and Fig.~\ref{fig:ab2} highlight the importance of HAR. Without HAR, the model can filter some false positives, but the transferred distribution remains largely constrained by the human prior, limiting its ability to relocate affordances to robot-feasible contacts. Adding HAR alone improves IoU$_{0.9}$ by $+27.1\%$ and F1$_{0.9}$ by $+27.5\%$ on the Unseen split, and by $+7.8\%$ and $+6.6\%$ on the Seen split, demonstrating that HAR's contribution is not contingent on prior noise but addresses the fundamental topological mismatch between human and robot affordances. While GRG filters geometric false positives that arise from embodiment differences, HAR performs the structural transformation: it leverages learnable robot prototypes to hierarchically reconstruct human semantics into robot-structured distributions, then applies graph diffusion to smooth the spatial topology consistent with robotic execution. The improvement concentrates on high-confidence metrics (IoU$_{0.9}$, F1$_{0.9}$) rather than threshold-averaged aIoU, indicating that HAR sharpens the robot-specific contact patterns rather than merely adjusting the overall distribution. Fig.~\ref{fig:ab2} illustrates this mechanism. The robustness of both modules to corruption of the human prior and to point-cloud noise and dropout is analyzed in Appendices~F and~G.

\subsubsection{\texorpdfstring{\textbf{Capacity versus Mechanisms}}{Capacity versus Mechanisms}}
\label{sec:capacity}
The gains of ART could in principle stem from model capacity or decoder complexity rather than from its transfer mechanisms. We test this with two controls. The PointNeXt backbones of FiLM and Cross-Attention are deepened until their parameters and GFLOPs match those of ART (FiLM-L and Cross-Attention-L), and HAR is replaced by a generic decoder of the same budget, built from self-attention and MLP blocks at three scales. Fig.~\ref{fig:capacity} plots accuracy against the number of parameters, with the full results given in Appendix~D. Matching the capacity of the baselines changes their accuracy only marginally, so the margin of ART over them is not a capacity effect. The generic decoder remains at the level of ART w/o HAR, so a decoder of equal budget does not substitute for the structure of HAR. Conversely, the ablation variants of Tab.~\ref{table:ablation} are of similar size yet span a wide range of accuracy that is produced by adding GRG and HAR. The gains of ART are therefore attributable to the two transfer mechanisms rather than to capacity or decoder complexity.

\begin{figure*}[t]
\centering
\includegraphics[width=\textwidth]{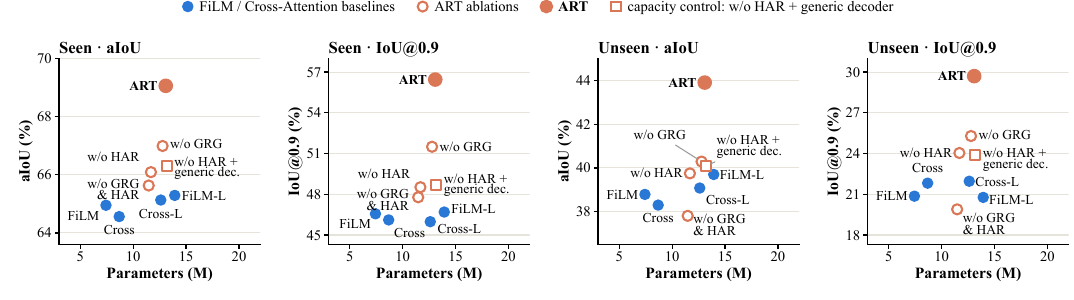}
\caption{\textbf{Accuracy versus number of parameters on HPAD.} aIoU and IoU$_{0.9}$ on Seen and Unseen for FiLM and Cross-Attention, their deepened variants FiLM-L and Cross-Attention-L, the ablation variants of ART, the generic-decoder control, and ART.}
\label{fig:capacity}
\end{figure*}

\subsection{Analytical Study}

\subsubsection{\texorpdfstring{\textbf{Action Injection Analysis}}{Action Injection Analysis}}
\begin{figure}[t]
    \centering
    \begin{overpic}[width=\columnwidth]{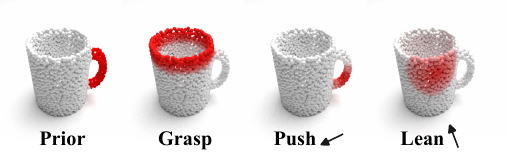}

    \end{overpic}
    \caption{\textbf{Action-Conditioned Transfer.} Visualizations of human prior and predicted affordances across different actions.}
    \label{fig:ab_mul}
    \vspace{-4mm}

\end{figure}

A core characteristic of HPAD is that a single object is paired with multiple atomic actions, where in some cases the human-prior regions for different actions overlap substantially even though the underlying robot affordance changes with each distinct contact pattern. ART resolves this one-to-many mapping through dual-path action injection: a global context vector encoding the action and object size modulates the backbone via AdaGN, and learnable action-indexed robot prototypes $\mathbf{Z}_a$ are injected into HAR through cross-attention. Fig.~\ref{fig:ab_mul} illustrates this: given the same object and a largely shared human prior, the predicted robot affordances shift markedly to action-appropriate contact configurations across actions.

\subsubsection{\texorpdfstring{\textbf{Robot-Data Efficiency}}{Robot-Data Efficiency}}
\label{sec:data_eff}
\begin{figure*}[t]
\centering
\captionsetup[subfigure]{labelfont=bf,textfont=normalfont}
\begin{subfigure}[b]{\textwidth}
\centering
\includegraphics[width=\linewidth]{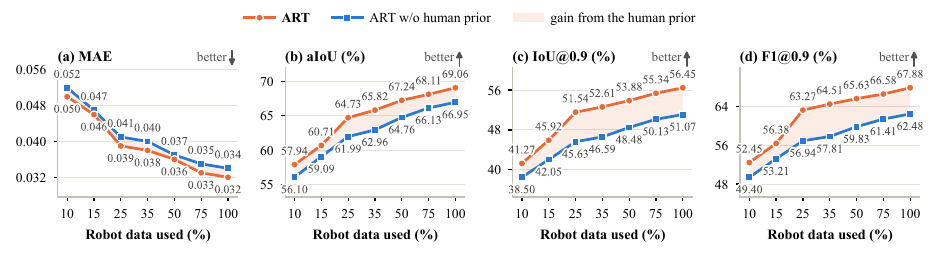}
\caption{\textbf{Seen categories.} ART leads ART w/o prior at every amount of robot data on all four metrics.}
\label{fig:data_eff_seen}
\end{subfigure}\\[4pt]
\begin{subfigure}[b]{\textwidth}
\centering
\includegraphics[width=\linewidth]{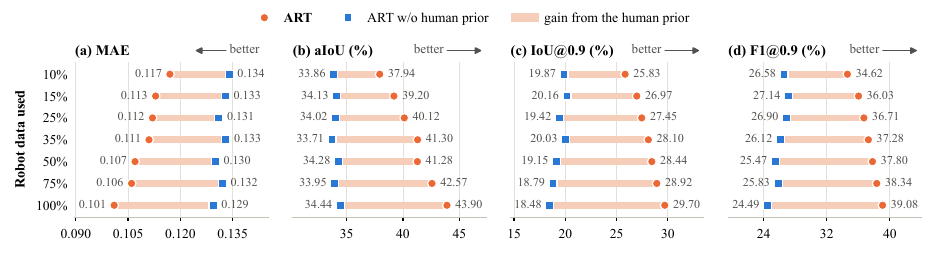}
\caption{\textbf{Unseen categories.} Accuracy stays flat without the prior, whereas with it ART is far more data-efficient.}
\label{fig:data_eff_unseen}
\end{subfigure}
\caption{\textbf{Robot-data efficiency.} MAE, aIoU, IoU$_{0.9}$ and F1$_{0.9}$ of ART and ART w/o prior versus the amount of robot data on Seen and Unseen categories. The shaded bands and the bars mark the gain from the human prior.}
\label{fig:data_eff}
\end{figure*}

Since ART is trained on robot-affordance labels, we examine how much robot data it requires and to what extent the human prior reduces this requirement. We retrain ART and ART w/o prior on nested subsets comprising 10\% to 75\% of the robot-labeled training objects of each split, subsampled within every category, and compare them with the models trained on the full set (Fig.~\ref{fig:data_eff}). The two splits address complementary questions. Seen tests whether the prior remains beneficial when robot data for the category is available, whereas Unseen corresponds to the practical case of novel categories for which no robot data is collected and is the more informative measure of data efficiency. On Seen, ART surpasses ART w/o prior at every amount of robot data on all four metrics, indicating that the prior remains beneficial even when robot data is abundant. On Unseen, additional robot data on the training categories does not transfer to novel categories without the prior, and ART w/o prior stays within one aIoU point of the Robot-Only PointNeXt baseline throughout. With the prior, 25\% of the robot labels suffice to exceed every baseline of Tab.~\ref{table:results} trained on the full set on aIoU, IoU$_{0.9}$ and F1$_{0.9}$, and accuracy improves further up to 100\%. The human prior therefore reduces the robot interaction data required, most substantially for categories without any robot data.

\begin{figure}[t]
\centering
\includegraphics[width=0.95\columnwidth]{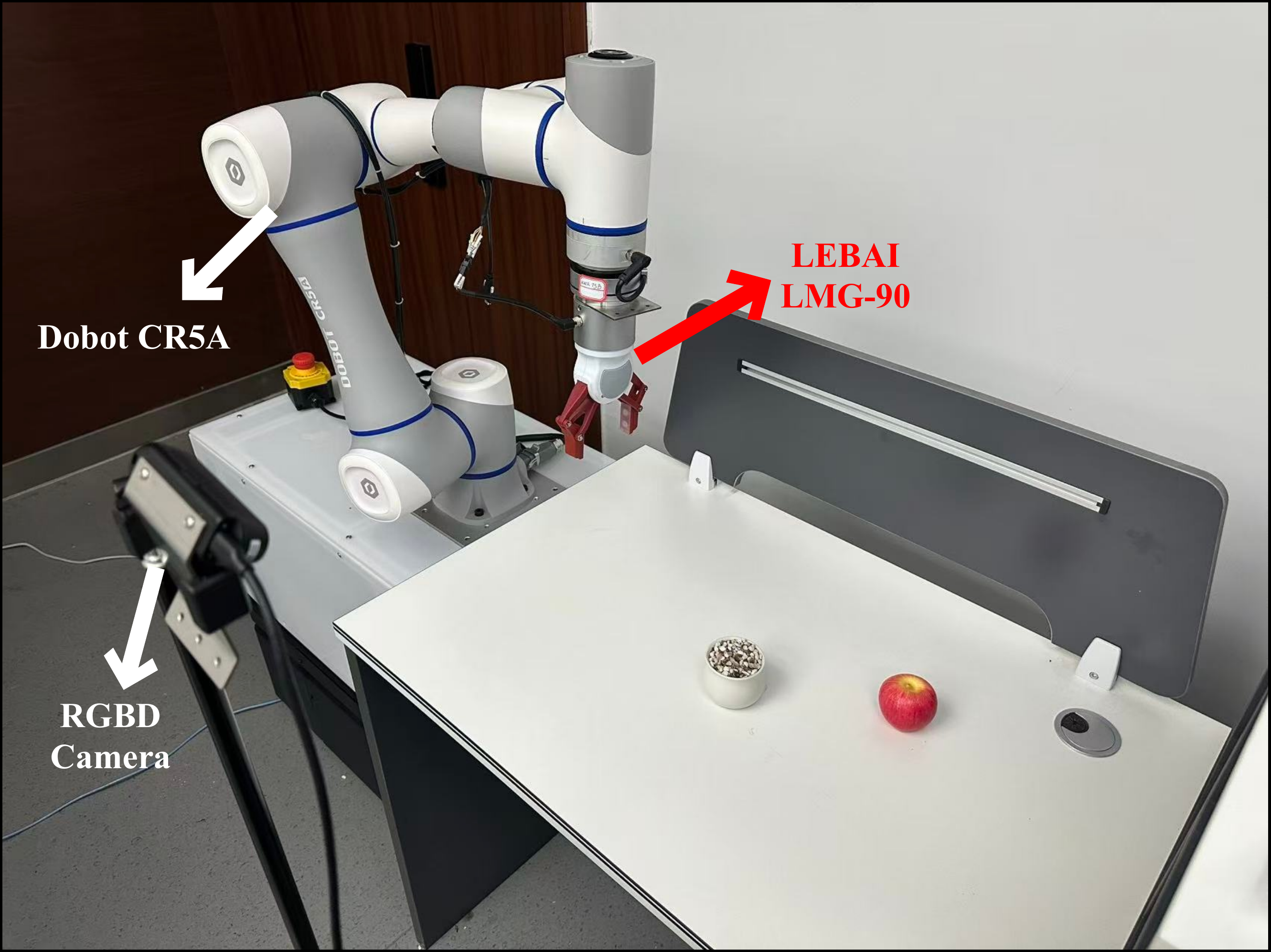}
\caption{\textbf{Real-world evaluation setup.} 6-DoF robot arm equipped with a parallel-jaw gripper. Object point clouds are reconstructed from real multi-view RGB-D observations.}
\label{fig:realworld_env}
\vspace{-6mm}
\end{figure}

\section{Real-World Validation}
\label{sec:realworld}

\subsection{Setup and Protocol}

\begin{table*}[t]
\centering
\caption{\textbf{Real-world manipulation success rates (\%).} Success rates are computed over 10 physical trials per (object, action, method) configuration for five methods: Human Prior (H-P), Cross-Attention (C-A), FiLM (F-M), ART w/o prior (A-N), and ART. \textbf{Bold} marks the best method per (object, action), and gray shading highlights ART entries. $\dagger$ denotes objects not present in HPAD.}
\label{tab:realworld}
\resizebox{\textwidth}{!}{
\setlength{\tabcolsep}{3pt}
\renewcommand{\arraystretch}{1.22}
\begin{tabular}{l|ccccc|ccccc|ccccc|ccccc}
\toprule
\textbf{Action} & \multicolumn{5}{c|}{\textbf{Grasp}}
 & \multicolumn{5}{c|}{\textbf{Push}}
 & \multicolumn{5}{c|}{\textbf{Drag}}
 & \multicolumn{5}{c}{\textbf{Lean}} \\
\midrule
\textbf{Object}
 & H-P & C-A & F-M & A-N & ART
 & H-P & C-A & F-M & A-N & ART
 & H-P & C-A & F-M & A-N & ART
 & H-P & C-A & F-M & A-N & ART \\
\midrule
Apple & 60 & 80 & 80 & \textbf{90} & \cellcolor{mygray}\textbf{90} & 80 & 90 & \textbf{100} & 80 & \cellcolor{mygray}90 & 60 & 80 & \textbf{90} & 80 & \cellcolor{mygray}\textbf{90} & 40 & 70 & 70 & 70 & \cellcolor{mygray}\textbf{80} \\
Cap & 20 & \textbf{70} & 60 & 60 & \cellcolor{mygray}\textbf{70} & 30 & 50 & 50 & 60 & \cellcolor{mygray}\textbf{70} & 70 & 50 & 60 & \textbf{80} & \cellcolor{mygray}\textbf{80} & 40 & 60 & 50 & 60 & \cellcolor{mygray}\textbf{70} \\
Flowerpot & 50 & 50 & 60 & 80 & \cellcolor{mygray}\textbf{90} & 90 & \textbf{100} & 80 & 90 & \cellcolor{mygray}\textbf{100} & 60 & 80 & \textbf{90} & 70 & \cellcolor{mygray}80 & 70 & 60 & 70 & 80 & \cellcolor{mygray}\textbf{90} \\
Doll & 70 & 60 & 80 & 50 & \cellcolor{mygray}\textbf{100} & 40 & 40 & 60 & 70 & \cellcolor{mygray}\textbf{80} & 50 & \textbf{90} & 80 & 70 & \cellcolor{mygray}70 & 50 & 70 & 70 & 70 & \cellcolor{mygray}\textbf{80} \\
Cup & 50 & 60 & 60 & 70 & \cellcolor{mygray}\textbf{80} & 80 & 90 & 90 & 90 & \cellcolor{mygray}\textbf{100} & 80 & 90 & 90 & \textbf{100} & \cellcolor{mygray}\textbf{100} & \textbf{60} & 40 & 50 & 20 & \cellcolor{mygray}50 \\
Bottle & 60 & 80 & 80 & 90 & \cellcolor{mygray}\textbf{100} & 60 & 80 & \textbf{90} & 80 & \cellcolor{mygray}\textbf{90} & 70 & 90 & \textbf{100} & \textbf{100} & \cellcolor{mygray}\textbf{100} & 40 & 80 & 70 & \textbf{90} & \cellcolor{mygray}\textbf{90} \\
Alarm Clock & 60 & 80 & \textbf{90} & 80 & \cellcolor{mygray}80 & 70 & \textbf{90} & 70 & 80 & \cellcolor{mygray}80 & 40 & 80 & \textbf{90} & 80 & \cellcolor{mygray}\textbf{90} & 40 & 70 & \textbf{80} & 70 & \cellcolor{mygray}70 \\
Headset & 70 & 80 & 70 & 80 & \cellcolor{mygray}\textbf{90} & 90 & \textbf{100} & 90 & \textbf{100} & \cellcolor{mygray}\textbf{100} & 60 & \textbf{80} & 60 & 70 & \cellcolor{mygray}70 & 60 & 80 & 90 & \textbf{100} & \cellcolor{mygray}\textbf{100} \\
Cube$^\dagger$ & 60 & 80 & 90 & 90 & \cellcolor{mygray}\textbf{100} & 80 & 70 & 80 & 60 & \cellcolor{mygray}\textbf{100} & 80 & \textbf{100} & \textbf{100} & 80 & \cellcolor{mygray}\textbf{100} & 60 & 80 & 90 & 50 & \cellcolor{mygray}\textbf{100} \\
Phone Stand$^\dagger$ & 40 & 60 & 50 & 60 & \cellcolor{mygray}\textbf{80} & 50 & 70 & 70 & 60 & \cellcolor{mygray}\textbf{100} & 70 & 60 & 80 & 50 & \cellcolor{mygray}\textbf{100} & 50 & 20 & 30 & 30 & \cellcolor{mygray}\textbf{60} \\
Coffee Tray$^\dagger$ & 70 & 80 & \textbf{90} & 80 & \cellcolor{mygray}\textbf{90} & 70 & 90 & 90 & 90 & \cellcolor{mygray}\textbf{100} & 70 & 60 & 50 & 70 & \cellcolor{mygray}\textbf{90} & 60 & 30 & 30 & 50 & \cellcolor{mygray}\textbf{70} \\
Mouse$^\dagger$ & 50 & 70 & 60 & 50 & \cellcolor{mygray}\textbf{80} & 60 & 60 & 70 & 50 & \cellcolor{mygray}\textbf{100} & 50 & 50 & 60 & 40 & \cellcolor{mygray}\textbf{80} & 10 & 40 & 50 & 30 & \cellcolor{mygray}\textbf{80} \\
\midrule
\textbf{Average} & 55.0 & 70.8 & 72.5 & 73.3 & \cellcolor{mygray}\textbf{87.5} & 66.7 & 77.5 & 78.3 & 75.8 & \cellcolor{mygray}\textbf{92.5} & 63.3 & 75.8 & 79.2 & 74.2 & \cellcolor{mygray}\textbf{87.5} & 48.3 & 58.3 & 62.5 & 60.0 & \cellcolor{mygray}\textbf{78.3} \\
\bottomrule
\end{tabular}
} 
\end{table*}

\begin{figure*}[t]
\centering
\includegraphics[width=\textwidth]{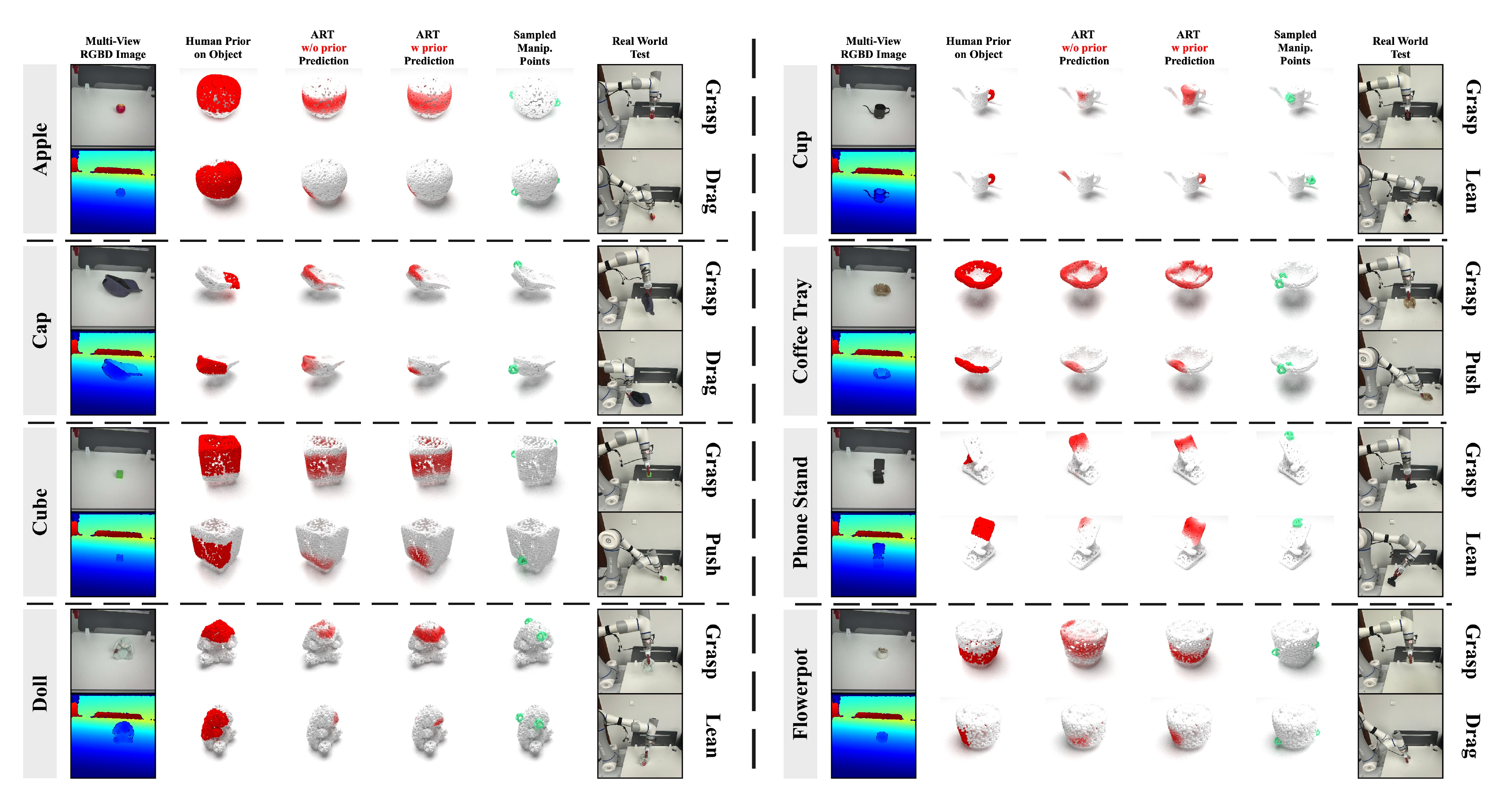}
\caption{\textbf{Real-world execution pipeline.} Each row shows one (object, action) example; from left to right, the columns are the RGB-D input, the human prior on the object, the affordance predicted by ART w/o prior, the transferred affordance produced by ART, the manipulation points sampled from the ART map, and the corresponding real-world execution by ART{}. \textcolor{red}{Red regions} indicate affordance maps, and \textcolor{green}{green points} represent sampled contact points.}
\label{fig:realworld_pipeline}
\end{figure*}

\textbf{Hardware and Protocol.} We evaluate ART on a physical Dobot CR5A 6-DoF robot arm equipped with a Lebai parallel-jaw gripper (Fig.~\ref{fig:realworld_env}). The benchmark covers $12$ objects spanning diverse geometry and materials, $4$ of which (marked $\dagger$) are not present in HPAD and serve as out-of-distribution probes. Each object is evaluated under $4$ action categories (Grasp, Push, Drag, and Lean), which group the 13 HPAD primitives by interaction family. We test five affordance methods (Human Prior, Cross-Attention, FiLM, ART w/o prior, and ART) with $10$ trials per (object, action category, method) configuration, totaling $2{,}400$ physical executions. Each object's point cloud is reconstructed from real multi-view RGB-D observations, so the input geometry reflects sensor noise and reconstruction artifacts. To execute an action, we sample candidate action points from the reconstructed cloud and rank them by the predicted affordance map. For Grasp, we sample antipodal grasps; for Push, Drag, and Lean, we sample surface contact points. The top-ranked feasible candidate is then executed (Fig.~\ref{fig:realworld_pipeline}, sampled points and execution). All methods share the same reconstructed point cloud, controller, point-selection rules, and success criteria; only the affordance map differs.

\subsection{Real-World Manipulation Results}

\subsubsection{\textbf{Overall Performance and Transfer Effectiveness}}
In this $2{,}400$-trial benchmark (Tab.~\ref{tab:realworld}), each method is evaluated over $480$ physical trials. Applying the human prior directly (H-P) succeeds in only $58.3\%$ of its trials, confirming that the embodiment gap between a human hand and a parallel-jaw gripper is substantial. The fusion baselines, Cross-Attention (C-A) and FiLM (F-M), improve markedly over the raw prior, indicating that learned fusion of the prior with robot-specific features bridges part of the gap. ART reaches $86.5\%$, well ahead of both baselines. This additional gain stems from its explicit treatment of the false positives in the prior, filtered by GRG, and of the distribution mismatch across embodiments, resolved by HAR. The same network without the prior, ART w/o prior (A-N), falls to $70.8\%$, the level of the fusion baselines. Its drop from ART appears in every action category but is concentrated on the four objects absent from HPAD, where the loss is nearly four times that on the eight HPAD objects. This matches the pattern observed on HPAD (Tab.~\ref{table:results}), where the prior contributes most on categories without robot labels. Further details of the real-world trials are given in Appendix~E.

\subsubsection{\textbf{Per-Action and Per-Object Analysis}}
Per-action results (Tab.~\ref{tab:realworld}) show that the advantage of ART is most pronounced for actions that depend on precise contact localization. Grasp requires antipodal contact pairs that ensure a stable closure, and Lean requires load-bearing surface patches that carry the object's weight during vertical motion. Both actions are therefore sensitive to the selected contact points and exhibit the largest margins over the fusion baselines. Drag, which tolerates contact-point error under horizontal tabletop-supported motion, yields the smallest margin. The contribution of the prior itself shows the same tendency, since the gain of ART over A-N is largest on Lean and smallest on Drag. Across objects, ART remains consistently high over diverse geometries and materials, including the four objects absent from HPAD. On these objects its margin over A-N is far wider than on the HPAD objects, so the transfer to unseen real objects rests on the human prior. Fig.~\ref{fig:realworld_pipeline} illustrates representative cases in which the transferred affordance maps lead to successful manipulation.

\section{Conclusion and Future Work}
\textbf{Conclusion.} We formulate a new setting for bridging embodiment gaps by transferring 3D human priors into robot-executable affordances. To support this setting, we introduce HPAD, a testbed of paired human priors and simulation-validated robot affordance annotations. We further propose a framework that explicitly addresses false positives in human priors and the distribution mismatch across embodiments. Extensive experiments confirm the effectiveness of the setting, and a 2{,}400-execution physical benchmark on a parallel-jaw 6-DoF platform shows ART attains the best success rate, 86.5\% over its 480 trials, supporting robot-actionable affordance transfer under real-world perception and execution conditions.

\textbf{Future Work.} We focus on a parallel-jaw gripper setting in this work, motivated by its prevalence in industrial deployment and by its substantial morphological and contact-mode gap from the human hand, which makes it a representative test case for human-to-robot affordance transfer. Different end-effector morphologies, in particular dexterous hands, exhibit qualitatively distinct contact primitives and target affordance distributions, so extending the proposed paradigm to these morphologies is a natural next direction. A further step is to make affordance transfer itself embodiment-conditioned, generalizing across arbitrary end-effectors without per-gripper specialization. We hope this work motivates future research on reliable and transferable 3D affordance learning.

\section*{Conflict of Interest}
The authors declare no conflicts of interest relevant to this work.

\bibliographystyle{IEEEtran}
\bibliography{references}

\begin{thebibliography}{10}
\providecommand{\url}[1]{#1}
\csname url@samestyle\endcsname
\providecommand{\newblock}{\relax}
\providecommand{\bibinfo}[2]{#2}
\providecommand{\BIBentrySTDinterwordspacing}{\spaceskip=0pt\relax}
\providecommand{\BIBentryALTinterwordstretchfactor}{4}
\providecommand{\BIBentryALTinterwordspacing}{\spaceskip=\fontdimen2\font plus
\BIBentryALTinterwordstretchfactor\fontdimen3\font minus \fontdimen4\font\relax}
\providecommand{\BIBforeignlanguage}[2]{{%
\expandafter\ifx\csname l@#1\endcsname\relax
\typeout{** WARNING: IEEEtran.bst: No hyphenation pattern has been}%
\typeout{** loaded for the language `#1'. Using the pattern for}%
\typeout{** the default language instead.}%
\else
\language=\csname l@#1\endcsname
\fi
#2}}
\providecommand{\BIBdecl}{\relax}
\BIBdecl

\bibitem{aff}
J.~J. Gibson, \emph{The ecological approach to visual perception: classic edition}.\hskip 1em plus 0.5em minus 0.4em\relax Psychology press, 2014.

\bibitem{intro_1_1}
T.~Nagarajan and K.~Grauman, ``Learning affordance landscapes for interaction exploration in 3d environments,'' \emph{Advances in Neural Information Processing Systems}, vol.~33, pp. 2005--2015, 2020.

\bibitem{intro_1_2}
P.~Mandikal and K.~Grauman, ``Learning dexterous grasping with object-centric visual affordances,'' in \emph{2021 IEEE international conference on robotics and automation (ICRA)}.\hskip 1em plus 0.5em minus 0.4em\relax IEEE, 2021, pp. 6169--6176.

\bibitem{eff}
J.~Borja-Diaz, O.~Mees, G.~Kalweit, L.~Hermann, J.~Boedecker, and W.~Burgard, ``Affordance learning from play for sample-efficient policy learning,'' in \emph{2022 International Conference on Robotics and Automation (ICRA)}.\hskip 1em plus 0.5em minus 0.4em\relax IEEE, 2022, pp. 6372--6378.

\bibitem{where2act}
K.~Mo, L.~J. Guibas, M.~Mukadam, A.~Gupta, and S.~Tulsiani, ``Where2act: From pixels to actions for articulated 3d objects,'' in \emph{Proceedings of the IEEE/CVF International Conference on Computer Vision}, 2021, pp. 6813--6823.

\bibitem{rlafford}
Y.~Geng, B.~An, H.~Geng, Y.~Chen, Y.~Yang, and H.~Dong, ``Rlafford: End-to-end affordance learning for robotic manipulation,'' in \emph{2023 IEEE International conference on robotics and automation (ICRA)}.\hskip 1em plus 0.5em minus 0.4em\relax IEEE, 2023, pp. 5880--5886.

\bibitem{afforddp}
S.~Wu, Y.~Zhu, Y.~Huang, K.~Zhu, J.~Gu, J.~Yu, Y.~Shi, and J.~Wang, ``Afforddp: Generalizable diffusion policy with transferable affordance,'' in \emph{Proceedings of the Computer Vision and Pattern Recognition Conference}, 2025, pp. 6971--6980.

\bibitem{AffordanceNet21}
S.~Deng, X.~Xu, C.~Wu, K.~Chen, and K.~Jia, ``3d affordancenet: A benchmark for visual object affordance understanding,'' in \emph{proceedings of the IEEE/CVF conference on computer vision and pattern recognition}, 2021, pp. 1778--1787.

\bibitem{nguyen2017}
A.~Nguyen, D.~Kanoulas, D.~G. Caldwell, and N.~G. Tsagarakis, ``Object-based affordances detection with convolutional neural networks and dense conditional random fields,'' in \emph{2017 IEEE/RSJ International Conference on Intelligent Robots and Systems (IROS)}, 2017, pp. 5908--5915.

\bibitem{luo2022}
H.~Luo, W.~Zhai, J.~Zhang, Y.~Cao, and D.~Tao, ``Learning affordance grounding from exocentric images,'' in \emph{Proceedings of the IEEE/CVF Conference on Computer Vision and Pattern Recognition}, 2022, pp. 2252--2261.

\bibitem{GREAT25}
Y.~Shao, W.~Zhai, Y.~Yang, H.~Luo, Y.~Cao, and Z.-J. Zha, ``Great: Geometry-intention collaborative inference for open-vocabulary 3d object affordance grounding,'' in \emph{Proceedings of the Computer Vision and Pattern Recognition Conference}, 2025, pp. 17\,326--17\,336.

\bibitem{AffordanceLLM3D25}
H.~Chu, X.~Deng, Q.~Lv, X.~Chen, Y.~Li, J.~Hao, and L.~Nie, ``3d-affordancellm: Harnessing large language models for open-vocabulary affordance detection in 3d worlds,'' \emph{arXiv preprint arXiv:2502.20041}, 2025.

\bibitem{lee2025}
J.~Lee, E.~Park, C.~Park, D.~Kang, and M.~Cho, ``Affogato: Open-vocabulary affordance grounding with automated data generation at scale,'' \emph{arXiv preprint arXiv:2506.12009}, 2025.

\bibitem{roboabc}
Y.~Ju, K.~Hu, G.~Zhang, G.~Zhang, M.~Jiang, and H.~Xu, ``Robo-abc: Affordance generalization beyond categories via semantic correspondence for robot manipulation,'' in \emph{European Conference on Computer Vision}.\hskip 1em plus 0.5em minus 0.4em\relax Springer, 2024, pp. 222--239.

\bibitem{ram}
Y.~Kuang, J.~Ye, H.~Geng, J.~Mao, C.~Deng, L.~Guibas, H.~Wang, and Y.~Wang, ``Ram: Retrieval-based affordance transfer for generalizable zero-shot robotic manipulation,'' \emph{arXiv preprint arXiv:2407.04689}, 2024.

\bibitem{IAGNet23}
Y.~Yang, W.~Zhai, H.~Luo, Y.~Cao, J.~Luo, and Z.-J. Zha, ``Grounding 3d object affordance from 2d interactions in images,'' in \emph{Proceedings of the IEEE/CVF International Conference on Computer Vision}, 2023, pp. 10\,905--10\,915.

\bibitem{MIFAG24}
X.~Gao, P.~Zhang, D.~Qu, D.~Wang, Z.~Wang, Y.~Ding, and B.~Zhao, ``Learning 2d invariant affordance knowledge for 3d affordance grounding,'' in \emph{Proceedings of the AAAI Conference on Artificial Intelligence}, vol.~39, no.~3, 2025, pp. 3095--3103.

\bibitem{DAViD25}
H.~Kim, S.~Baik, and H.~Joo, ``David: Modeling dynamic affordance of 3d objects using pre-trained video diffusion models,'' in \emph{Proceedings of the IEEE/CVF International Conference on Computer Vision}, 2025, pp. 10\,330--10\,341.

\bibitem{UAD25}
Y.~Tang, W.~Huang, Y.~Wang, C.~Li, R.~Yuan, R.~Zhang, J.~Wu, and L.~Fei-Fei, ``Uad: Unsupervised affordance distillation for generalization in robotic manipulation,'' \emph{arXiv preprint arXiv:2506.09284}, 2025.

\bibitem{lu2023phrase}
L.~Lu, W.~Zhai, H.~Luo, Y.~Kang, and Y.~Cao, ``Phrase-based affordance detection via cyclic bilateral interaction,'' \emph{IEEE Transactions on Artificial Intelligence}, vol.~4, no.~5, pp. 1186--1198, 2023.

\bibitem{LangCondAffPose24}
T.~Nguyen, M.~N. Vu, B.~Huang, T.~Van~Vo, V.~Truong, N.~Le, T.~Vo, B.~Le, and A.~Nguyen, ``Language-conditioned affordance-pose detection in 3d point clouds,'' in \emph{2024 IEEE International Conference on Robotics and Automation (ICRA)}.\hskip 1em plus 0.5em minus 0.4em\relax IEEE, 2024, pp. 3071--3078.

\bibitem{LASO24}
Y.~Li, N.~Zhao, J.~Xiao, C.~Feng, X.~Wang, and T.-s. Chua, ``Laso: Language-guided affordance segmentation on 3d object,'' in \emph{Proceedings of the IEEE/CVF Conference on Computer Vision and Pattern Recognition}, 2024, pp. 14\,251--14\,260.

\bibitem{OpenAD23}
T.~Nguyen, M.~N. Vu, A.~Vuong, D.~Nguyen, T.~Vo, N.~Le, and A.~Nguyen, ``Open-vocabulary affordance detection in 3d point clouds,'' in \emph{2023 IEEE/RSJ International Conference on Intelligent Robots and Systems (IROS)}.\hskip 1em plus 0.5em minus 0.4em\relax IEEE, 2023, pp. 5692--5698.

\bibitem{AnyPart24}
T.~van Oort, D.~Miller, W.~N. Browne, N.~Marticorena, J.~Haviland, and N.~Suenderhauf, ``Open-vocabulary part-based grasping,'' \emph{arXiv preprint arXiv:2406.05951}, 2024.

\bibitem{OOAL24}
G.~Li, D.~Sun, L.~Sevilla-Lara, and V.~Jampani, ``One-shot open affordance learning with foundation models,'' in \emph{Proceedings of the IEEE/CVF Conference on Computer Vision and Pattern Recognition}, 2024, pp. 3086--3096.

\bibitem{AffordanceLLM24}
S.~Qian, W.~Chen, M.~Bai, X.~Zhou, Z.~Tu, and L.~E. Li, ``{AffordanceLLM}: Grounding affordance from vision language models,'' in \emph{Proceedings of the IEEE/CVF Conference on Computer Vision and Pattern Recognition (CVPR) Workshops}, 2024, pp. 7587--7597.

\bibitem{roy2026mutual}
P.~Roy, S.~Bhattacharya, S.~Ghosh, U.~Pal, and M.~Blumenstein, ``Exploring mutual cross-modal attention for context-aware human affordance generation,'' \emph{IEEE Transactions on Artificial Intelligence}, vol.~7, no.~2, pp. 806--817, 2026.

\bibitem{PAVLM24}
S.-C. Liu, V.~N. Tran, W.~Chen, W.-L. Cheng, Y.-L. Huang, I.-B. Liao, Y.-H. Li, and J.~Zhang, ``Pavlm: Advancing point cloud based affordance understanding via vision-language model,'' \emph{arXiv preprint arXiv:2410.11564}, 2024.

\bibitem{SeqAfford25}
C.~Yu, H.~Wang, Y.~Shi, H.~Luo, S.~Yang, J.~Yu, and J.~Wang, ``Seqafford: Sequential 3d affordance reasoning via multimodal large language model,'' in \emph{Proceedings of the Computer Vision and Pattern Recognition Conference}, 2025, pp. 1691--1701.

\bibitem{LangInstrAfford25}
H.~Zhu, Q.~Kong, K.~Xu, X.~Xia, B.~Deng, J.~Ye, R.~Xiong, and Y.~Wang, ``Grounding 3d object affordance with language instructions, visual observations and interactions,'' in \emph{Proceedings of the Computer Vision and Pattern Recognition Conference}, 2025, pp. 17\,337--17\,346.

\bibitem{GLOVER24}
T.~Ma, Z.~Wang, J.~Zhou, M.~Wang, and J.~Liang, ``Glover: Generalizable open-vocabulary affordance reasoning for task-oriented grasping,'' \emph{arXiv preprint arXiv:2411.12286}, 2024.

\bibitem{DAG25}
H.~Wang, Z.~Zhang, K.~Ji, M.~Liu, W.~Yin, Y.~Chen, Z.~Liu, X.~Zeng, T.~Gui, and H.~Zhang, ``{DAG}: Unleash the potential of diffusion model for open-vocabulary 3d affordance grounding,'' \emph{arXiv preprint arXiv:2508.01651}, 2025.

\bibitem{enayati2024sim2real}
A.~M.~S. Enayati, R.~Dershan, Z.~Zhang, D.~Richert, and H.~Najjaran, ``Facilitating sim-to-real by intrinsic stochasticity of real-time simulation in reinforcement learning for robot manipulation,'' \emph{IEEE Transactions on Artificial Intelligence}, vol.~5, no.~4, pp. 1791--1804, 2024.

\bibitem{fan2024safe}
K.~Fan, Z.~Chen, G.~Ferrigno, and E.~De~Momi, ``Learn from safe experience: Safe reinforcement learning for task automation of surgical robot,'' \emph{IEEE Transactions on Artificial Intelligence}, vol.~5, no.~7, pp. 3374--3383, 2024.

\bibitem{vrb}
S.~Bahl, R.~Mendonca, L.~Chen, U.~Jain, and D.~Pathak, ``Affordances from human videos as a versatile representation for robotics,'' in \emph{Proceedings of the IEEE/CVF Conference on Computer Vision and Pattern Recognition}, 2023, pp. 13\,778--13\,790.

\bibitem{hrp}
M.~K. Srirama, S.~Dasari, S.~Bahl, and A.~Gupta, ``Hrp: Human affordances for robotic pre-training,'' \emph{arXiv preprint arXiv:2407.18911}, 2024.

\bibitem{glover++}
T.~Ma, J.~Zheng, Z.~Wang, Z.~Gao, J.~Zhou, and J.~Liang, ``Glover++: Unleashing the potential of affordance learning from human behaviors for robotic manipulation,'' \emph{arXiv preprint arXiv:2505.11865}, 2025.

\bibitem{2handedafforder}
M.~Heidinger, S.~Jauhri, V.~Prasad, and G.~Chalvatzaki, ``2handedafforder: Learning precise actionable bimanual affordances from human videos,'' \emph{arXiv preprint arXiv:2503.09320}, 2025.

\bibitem{generalflow}
C.~Yuan, C.~Wen, T.~Zhang, and Y.~Gao, ``General flow as foundation affordance for scalable robot learning,'' \emph{arXiv preprint arXiv:2401.11439}, 2024.

\bibitem{oval}
E.~Tong, A.~Opipari, S.~Lewis, Z.~Zeng, and O.~C. Jenkins, ``Oval-prompt: Open-vocabulary affordance localization for robot manipulation through llm affordance-grounding,'' \emph{arXiv preprint arXiv:2404.11000}, 2024.

\bibitem{pointnext}
G.~Qian, Y.~Li, H.~Peng, J.~Mai, H.~Hammoud, M.~Elhoseiny, and B.~Ghanem, ``Pointnext: Revisiting pointnet++ with improved training and scaling strategies,'' \emph{Advances in neural information processing systems}, vol.~35, pp. 23\,192--23\,204, 2022.

\bibitem{dhariwal2021adagn}
P.~Dhariwal and A.~Q. Nichol, ``Diffusion models beat {GAN}s on image synthesis,'' in \emph{Advances in Neural Information Processing Systems}, 2021.

\bibitem{objaverse}
M.~Deitke, D.~Schwenk, J.~Salvador, L.~Weihs, O.~Michel, E.~VanderBilt, L.~Schmidt, K.~Ehsani, A.~Kembhavi, and A.~Farhadi, ``Objaverse: A universe of annotated 3d objects,'' in \emph{Proceedings of the IEEE/CVF conference on computer vision and pattern recognition}, 2023, pp. 13\,142--13\,153.

\bibitem{robosuite}
Y.~Zhu, J.~Wong, A.~Mandlekar, R.~Mart{\'\i}n-Mart{\'\i}n, A.~Joshi, S.~Nasiriany, and Y.~Zhu, ``robosuite: A modular simulation framework and benchmark for robot learning,'' \emph{arXiv preprint arXiv:2009.12293}, 2020.

\bibitem{mimicgen}
A.~Mandlekar, S.~Nasiriany, B.~Wen, I.~Akinola, Y.~Narang, L.~Fan, Y.~Zhu, and D.~Fox, ``Mimicgen: A data generation system for scalable robot learning using human demonstrations,'' \emph{arXiv preprint arXiv:2310.17596}, 2023.

\bibitem{jian2023affordpose}
J.~Jian, X.~Liu, M.~Li, R.~Hu, and J.~Liu, ``Affordpose: A large-scale dataset of hand-object interactions with affordance-driven hand pose,'' in \emph{Proceedings of the IEEE/CVF International Conference on Computer Vision}, 2023.

\bibitem{geng2023gapartnet}
H.~Geng, H.~Xu, C.~Zhao, C.~Xu, L.~Yi, S.~Huang, and H.~Wang, ``Gapartnet: Cross-category domain-generalizable object perception and manipulation via generalizable and actionable parts,'' in \emph{Proceedings of the IEEE/CVF Conference on Computer Vision and Pattern Recognition}, 2023.

\bibitem{pointnet2}
C.~R. Qi, L.~Yi, H.~Su, and L.~J. Guibas, ``Pointnet++: Deep hierarchical feature learning on point sets in a metric space,'' \emph{Advances in neural information processing systems}, vol.~30, 2017.

\bibitem{attention}
A.~Vaswani, N.~Shazeer, N.~Parmar, J.~Uszkoreit, L.~Jones, A.~N. Gomez, {\L}.~Kaiser, and I.~Polosukhin, ``Attention is all you need,'' \emph{Advances in neural information processing systems}, vol.~30, 2017.

\bibitem{film}
E.~Perez, F.~Strub, H.~De~Vries, V.~Dumoulin, and A.~Courville, ``Film: Visual reasoning with a general conditioning layer,'' in \emph{Proceedings of the AAAI conference on artificial intelligence}, vol.~32, no.~1, 2018.

\end{thebibliography}

\end{document}